\documentclass{article}
\usepackage[main,final]{neurips_2026}

\usepackage[utf8]{inputenc} % allow utf-8 input
\usepackage[T1]{fontenc}    % use 8-bit T1 fonts
\usepackage{hyperref}       % hyperlinks
\usepackage{url}            % simple URL typesetting
\usepackage{booktabs}       % professional-quality tables
\usepackage{amsfonts}       % blackboard math symbols
\usepackage{nicefrac}       % compact symbols for 1/2, etc.
\usepackage{microtype}      % microtypography
\usepackage{xcolor}         % colors
\usepackage{amsmath}
\usepackage{graphicx}
\usepackage{multirow}
\newcommand{\uniformtablescale}{0.81}
\usepackage{circledsteps}
\title{MoCAR: Motion-code Coordinate-aware AutoRegression for Continuous Trajectory Forecasting}

\author{%
  \textbf{Yiming Xu}$^{1}$\thanks{Corresponding author: \texttt{Yiming.Xu@ikg.uni-hannover.de}.}
  \quad \textbf{Hao Cheng}$^{2}$ \quad \textbf{Monika Sester}$^{1}$ \\
  $^{1}$Institute of Cartography and Geoinformatics \\
  Leibniz University Hannover, 30167 Hannover, Germany \\
  $^{2}$Faculty of Geo-Information Science and Earth Observation \\
  University of Twente, 7522 NH Enschede, The Netherlands \\
  \texttt{\{Yiming.Xu,Monika.Sester\}@ikg.uni-hannover.de} \\
  \texttt{h.cheng-2@utwente.nl}
}

\begin{document}

\maketitle
\begin{figure}[ht]
    \centering
    % \vspace{3mm}
    \includegraphics[width=1\linewidth]{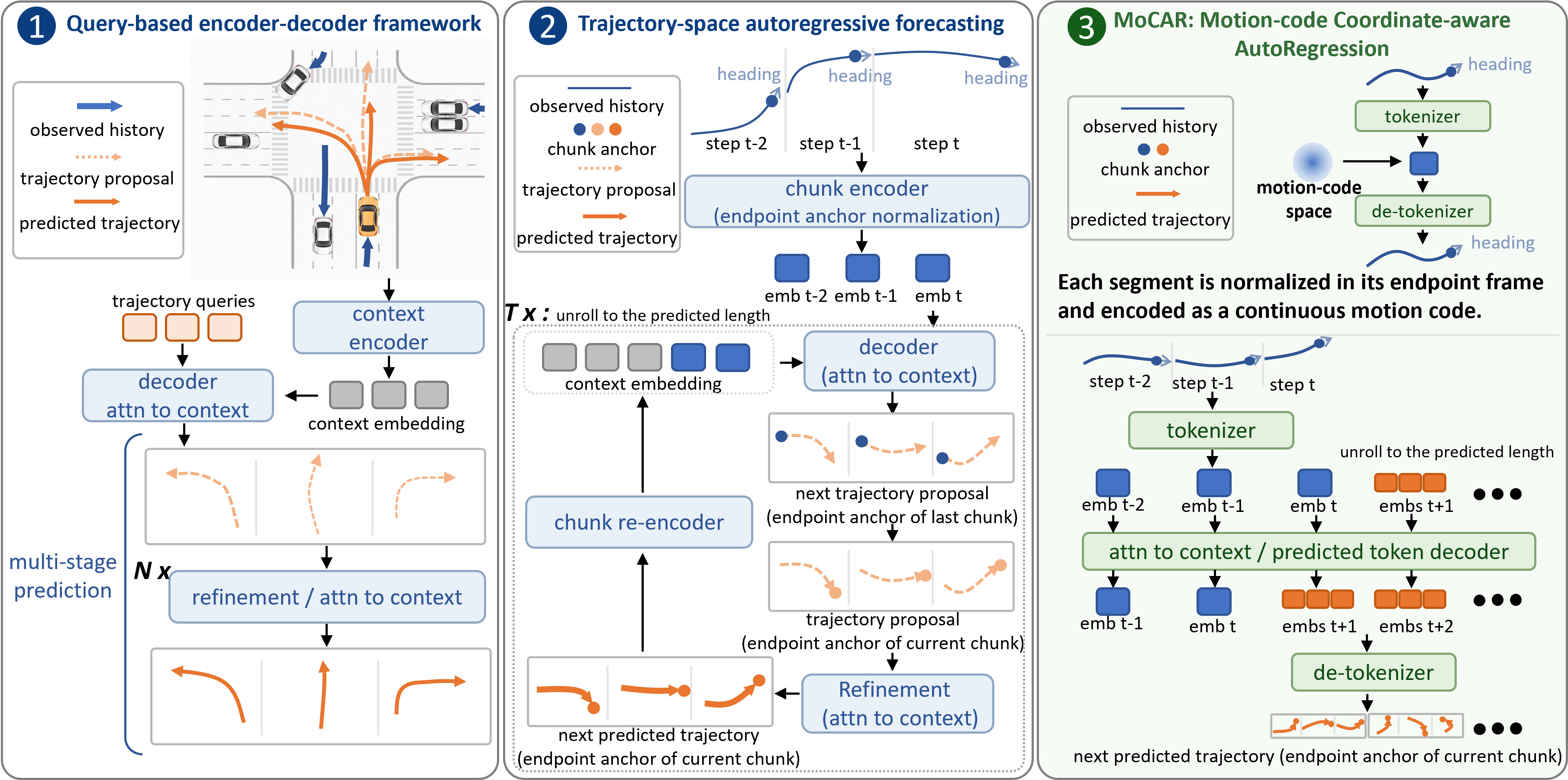}
    \caption{
    \textbf{Paradigm comparison.}
    \Circled{1} Query-based encoder--decoder methods generate trajectory proposals and refine them through additional decoding stages.
    \Circled{2} Trajectory-space autoregressive methods explicitly decode each sub-trajectory, update the local frame, and re-tokenize the predicted motion for the next step.
    \Circled{3} In contrast, \textbf{MoCAR} represents endpoint-normalized trajectory segments as continuous motion codes and performs autoregressive next-code prediction directly in latent space, avoiding explicit trajectory queries and proposal-and-refinement pipelines.
    }
    \label{fig:paradigm}
\end{figure}

\begin{abstract}
Autoregressive generation is natural for language, where predicted tokens can be directly reused as the next prediction state, but trajectory forecasting lacks such a clean token: motion is continuous, multimodal, and expressed in local coordinate frames that evolve with the predicted trajectory. We present \textbf{MoCAR} (\textbf{Mo}tion-code \textbf{C}oordinate-aware \textbf{A}uto\textbf{R}egression), a \emph{decoder-only} framework that casts trajectory forecasting as next-code prediction in a coordinate-aware continuous latent space. MoCAR learns a continuous motion-code space for endpoint-normalized trajectory segments, where each code jointly captures local trajectory geometry and the reference-frame transition induced by that segment. Historical motion codes are used as a teacher-forced prefix, future codes are generated autoregressively under temporal, map, agent, and mode interactions, and predicted codes persist in latent memory while decoded endpoints update the local scene context. This enables rollout without trajectory-space re-tokenization, trajectory queries, goal candidates, or proposal-and-refinement pipelines. On Argoverse (AV) benchmarks, MoCAR achieves top-tier performance with a simple single-stage architecture, transfers strongly from AV2 to AV1 in zero-shot evaluation, and improves on turn-heavy scenarios. Ablations confirm that the learned continuous motion-code space, latent alignment, weak KL regularization, and joint tokenizer--predictor optimization are essential for stable latent autoregression.
\end{abstract}

\section{Introduction}

Trajectory forecasting predicts agents' future motion from their observed histories, road context, and interactions with other agents. As an agent moves, the relevant lanes and neighboring agents change. Autoregressive forecasting can update this context after each predicted segment~\citep{knoche2025donut}. The key design choice is what motion representation to pass to the next prediction step. We propose \textbf{MoCAR} (\textbf{Mo}tion-code \textbf{C}oordinate-aware \textbf{A}uto\textbf{R}egression) to learn continuous motion codes for this purpose. Each code captures local trajectory geometry and the reference-frame transition induced by the segment. Predicted codes directly condition subsequent steps, while decoded endpoints update the local scene context. We refer to this direct reuse of the prediction state as \emph{clean autoregression}.

Learning a reusable motion unit poses three challenges. First, position, velocity, heading, and curvature vary continuously. The representation must preserve the small differences between nearby motions. Second, the same observed history can lead to different futures, such as going straight, turning, or yielding~\citep{shi2022motion,gao2020vectornet,zhou2023query}. Third, each predicted segment changes the position and heading used to query nearby lanes and agents~\citep{zhou2023query,knoche2025donut}. Unlike visual tokens that tile a fixed image grid, a motion token represents a \emph{state transition}. It must describe local trajectory geometry and the relative pose change that establishes the next frame. MoCAR addresses these requirements with coordinate-aware continuous codes and separate rollout branches for multiple possible futures.

Existing forecasting methods use different prediction units and decoding pipelines (Figure~\ref{fig:paradigm}). Vectorized and query-based encoder--decoder methods achieve strong forecasting performance~\citep{gao2020vectornet,zhou2022hivt,jia2023hdgt,nayakanti2023wayformer,zhou2023query}. Many use goal candidates, motion queries, anchor-informed proposals, or refinement stages~\citep{shi2022motion,wang2023prophnet,choi2023r,zhou2024smartrefine,zhang2024decoupling}. DONUT~\citep{knoche2025donut} unifies history encoding and future rollout in a decoder-only model. It updates the local frame along predicted segments, but each step still generates, refines, and re-embeds a trajectory segment. Proposal generation, refinement, and feedback re-encoding introduce additional computation. Task-specific decoding designs can also limit the reuse of learned motion representations across datasets and prediction horizons. MoCAR instead learns continuous motion codes that jointly represent local trajectory shape and the induced frame transition. Each predicted code directly conditions the next prediction, while decoding supplies the trajectory and the next scene anchor. Forecasting therefore follows a shared next-code prediction rule over reusable motion units, without feedback re-encoding or additional refinement.

We learn the motion codes with a variational autoencoder~\citep{kingma2013auto}. Each trajectory segment is normalized using the position and heading of its endpoint. The code thus captures local motion geometry and the reference-frame transition induced by the segment. Following DONUT's history-conditioning strategy~\citep{knoche2025donut}, codes from observed segments form a teacher-forced prefix. A shared causal predictor generates future codes using temporal memory, map context, and interactions among agents and modes. Learned mode embeddings initialize $K$ rollout branches for different future hypotheses. The tokenizer and de-tokenizer are pretrained for reconstruction and then jointly optimized with the forecaster. Weak KL regularization promotes a smooth code space, while latent alignment supervises predicted codes. This yields a single-stage decoder-only forecaster.

The learned code space also provides a reusable vocabulary of local state transitions. Endpoint normalization removes dependence on global position and orientation. This is useful for cross-dataset transfer, where observation and prediction horizons, coordinate conventions, and motion distributions can change~\citep{feng2024unitraj,yao2024improving}. On Argoverse (AV) benchmarks~\citep{wilson2023argoverse,chang2019argoverse}, MoCAR achieves top-tier AV2 accuracy, strong AV2-to-AV1 zero-shot transfer, and improved performance on turn-heavy scenes. Controlled ablations examine code learning and direct code feedback. A matched-stride comparison also shows lower latency than the strong baseline DONUT.

Our contributions are summarized as follows:
\begin{itemize}
    \item We introduce coordinate-aware continuous motion codes for trajectory segments. Each code represents local motion geometry and the reference-frame transition needed for the next scene update.
    \item We formulate trajectory forecasting as persistent latent autoregression over these codes. To our knowledge, MoCAR is the first decoder-only trajectory forecaster to directly predict and reuse such codes. It requires no trajectory-space re-tokenization or additional refinement.
    \item We demonstrate strong AV2 performance, AV2-to-AV1 zero-shot transfer, and gains on turn-heavy scenes. Controlled ablations examine the contributions of motion-code learning and persistent feedback. Matched-stride evaluation shows lower latency than the strong decoder-only baseline DONUT.
\end{itemize}

Our code is publicly available at \url{https://github.com/XuYiMing83/MoCAR}.

\section{Related Work}

\paragraph{Vectorized and query-based motion forecasting.}
Motion forecasting for autonomous driving has increasingly adopted vectorized scene representations. VectorNet~\citep{gao2020vectornet}, HiVT~\citep{zhou2022hivt}, HDGT~\citep{jia2023hdgt}, Wayformer~\citep{nayakanti2023wayformer}, and QCNet~\citep{zhou2023query} model map-agent interactions using polylines, graphs, and attention~\citep{vaswani2017attention}. Query-based methods introduce intention localization~\citep{shi2022motion,shi2024mtr++}, anchor-informed proposals~\citep{wang2023prophnet}, and decomposed query designs~\citep{zhang2024decoupling}. Other approaches refine trajectory predictions with tube queries~\citep{choi2023r} or adapt refinement to the scene~\citep{zhou2024smartrefine}. These methods use specialized queries, proposals, or refinement modules to generate future trajectories. MoCAR learns reusable local motion units and predicts them sequentially with a shared causal model. Each predicted unit directly conditions the next step as the scene context is updated.

\paragraph{Autoregressive motion forecasting and simulation.}
MotionLM~\citep{seff2023motionlm} and Trajeglish~\citep{philion2023trajeglish} model traffic through autoregressive prediction over discrete motion tokens. Both condition their motion decoders on encoded scene context. SMART~\citep{wu2024smart} uses discrete motion tokens with a decoder-only transformer for multi-agent simulation. BehaviorGPT~\citep{zhou2024behaviorgpt} uses a homogeneous autoregressive transformer and predicts trajectory patches. DONUT~\citep{knoche2025donut} unifies history encoding and future rollout in a decoder-only forecasting model. Its proposer--refiner module generates trajectory segments. These segments update the local reference frame and are re-embedded as input for the next step. MoCAR directly predicts continuous motion codes and reuses them as the feedback state. Decoded endpoints provide the reference poses for subsequent scene updates. The rollout uses no trajectory-space re-tokenization or additional refinement.

\paragraph{Motion tokenization and latent motion representations.}
Generative trajectory models include VAE-based latent-variable models~\citep{kingma2013auto,xu2022socialvae,xiang2024socialcvae}. Other approaches use diffusion~\citep{jiang2023motiondiffuser,xu2024controllable}, normalizing flows~\citep{liang2023stglow}, or flow matching~\citep{yan2025trajflow}. Many of these models generate a complete future trajectory or use a latent variable for the full prediction horizon. Discrete tokenization offers another way to represent motion~\citep{van2017neural,philion2023trajeglish}. MoCAR learns continuous VAE-based codes for local trajectory segments. Each segment is normalized using the position and heading of its endpoint. Its code captures both local trajectory geometry and the reference-frame transition needed for the next scene update. This continuous representation is learned jointly with the predictor and directly reused during autoregressive rollout.

\section{Method}
\label{sec:method}

\subsection{Problem formulation}

We consider a vectorized driving scene with $A$ agents and $M$ map elements. At timestep $t$, agent $i$ has state $\mathbf{x}_i^t=(\mathbf{p}_i^t,\mathbf{v}_i^t,\theta_i^t,\gamma_i)$. Here, $\mathbf{p}_i^t\in\mathbb{R}^2$ and $\mathbf{v}_i^t\in\mathbb{R}^2$ denote position and velocity. The heading is $\theta_i^t$, and $\gamma_i$ denotes the agent category. The observed history is $\mathbf{X}_i=\{\mathbf{x}_i^t\}_{t=1}^{T_h}$. The map contains vectorized elements $\mathcal{M}=\{L_j\}_{j=1}^{M}$. Each element contains sampled points and semantic attributes such as lane type or boundary type.

Given $\{\mathbf{X}_i\}_{i=1}^{A}$ and $\mathcal{M}$, the model predicts $K$ future trajectories $\hat{\mathbf{Y}}_i^{(k)}=\{(\hat{\mathbf{p}}_i^{t,k},\hat{\theta}_i^{t,k})\}_{t=T_h+1}^{T_h+T_f}$. Their probabilities $\{\pi_i^{(k)}\}_{k=1}^{K}$ satisfy $\sum_{k=1}^{K}\pi_i^{(k)}=1$. We represent motion as a sequence of local segments and forecast by autoregressive next-code prediction in a continuous latent space.

\subsection{Coordinate-aware continuous motion codes}

\begin{figure}[t]
    \centering
    \includegraphics[width=1\linewidth]{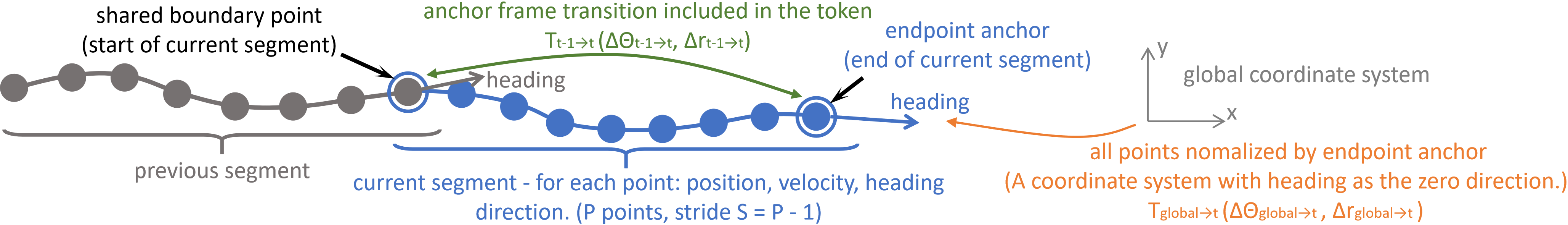}
    \caption{
    \textbf{Coordinate-aware motion segment.}
    Each local trajectory segment is normalized in the coordinate frame of its endpoint anchor and encoded as a continuous motion code.
    This code jointly represents local geometry and the induced frame transition, making it the autoregressive prediction unit in MoCAR.
    }
    \label{fig:motion_segment}
\end{figure}

We split each trajectory into segments of $P$ points with stride $S=P-1$. For segment index $c$, $\tau_c$ denotes the endpoint timestep. The segment of agent $i$ is $\mathcal{S}_{i,c}=\{(\mathbf{p}_i^{\tau_c-S+j},\theta_i^{\tau_c-S+j},\mathbf{v}_i^{\tau_c-S+j})\}_{j=0}^{P-1}$. Adjacent segments share one boundary point, as shown in Figure~\ref{fig:motion_segment}. Within each segment, we derive transition features including displacement, velocity direction, and heading change.

Each segment is represented in the local coordinate frame of its endpoint anchor $\mathbf{r}_{i,c}=(\mathbf{p}_i^{\tau_c},\theta_i^{\tau_c})$:
\begin{equation}
    \tilde{\mathbf{p}}_{i,c}^{j}=R(-\theta_i^{\tau_c})(\mathbf{p}_i^{\tau_c-S+j}-\mathbf{p}_i^{\tau_c}), \quad \tilde{\theta}_{i,c}^{j}=\mathrm{wrap}(\theta_i^{\tau_c-S+j}-\theta_i^{\tau_c}).
\end{equation}
Here, $R(\cdot)$ denotes the 2D rotation matrix and $\mathrm{wrap}(\cdot)$ maps angles to $[-\pi,\pi)$. The endpoint lies at the origin, and the local $x$-axis follows its heading. The normalized segment retains its start pose relative to the endpoint. This relative pose specifies the translation and rotation between consecutive endpoint frames. The motion code therefore jointly represents local trajectory shape and the associated frame transition. This allows the same local motion pattern to be reused at different scene locations and orientations.

We learn the motion-code space with a variational autoencoder. The encoder maps a normalized segment to a Gaussian distribution $q_{\phi}(\mathbf{z}_{i,c}\mid \tilde{\mathcal{S}}_{i,c})=\mathcal{N}(\boldsymbol{\mu}_{i,c},\mathrm{diag}(\boldsymbol{\sigma}_{i,c}^{2}))$. The decoder reconstructs the local segment as $\hat{\mathcal{S}}_{i,c}=D_{\psi}(\mathbf{z}_{i,c})$. We call $\mathbf{z}_{i,c}$ a \emph{motion code}, or \emph{token}. The \emph{motion-code space} comprises the tokenizer, de-tokenizer, and their continuous latent representation. We use the deterministic encoder mean $\boldsymbol{\mu}_{i,c}$ for historical conditioning and ground-truth future targets in latent supervision.

Reconstruction pretraining teaches the tokenizer and de-tokenizer to preserve local motion geometry and frame transitions. During forecasting, the decoder receives predicted codes, which may differ from those produced by the encoder. Latent alignment supervises these predictions toward the encoded future segments, while weak KL regularization encourages a smooth motion-code space. Joint optimization then adapts this representation to forecasting. A smaller learning rate for the tokenizer and de-tokenizer helps preserve their learned reconstruction structure.

\subsection{Decoder-only motion-code autoregression}
\label{sec:code_autoreg}
\begin{figure}[t]
    \centering
    \includegraphics[width=\linewidth]{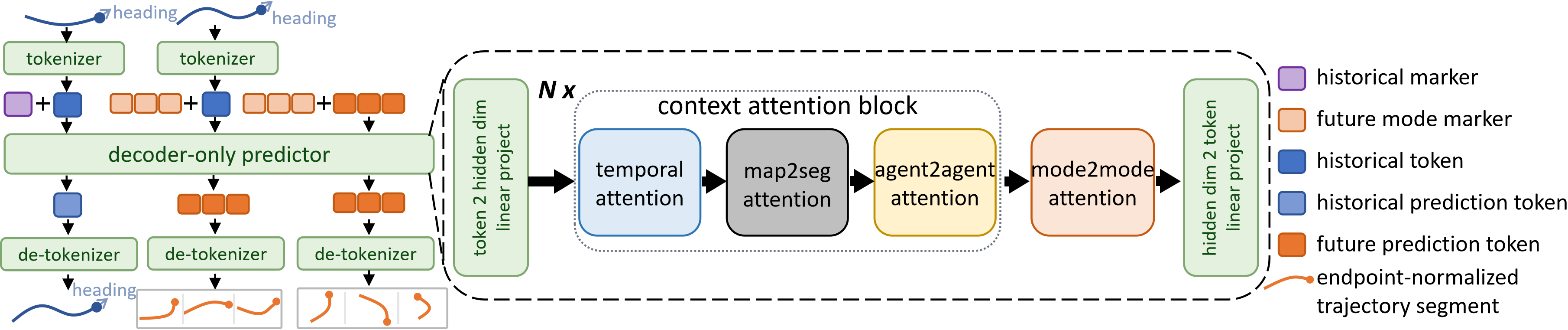}
    \caption{
    \textbf{Decoder-only motion-code autoregression in MoCAR.}
    Historical segments are tokenized into a teacher-forced motion-code prefix, and future prediction rolls out autoregressively in latent space.
    Mode embeddings separate latent rollout branches, while the predictor updates each branch using temporal, map, agent, and mode interactions before decoding the next segment and endpoint anchor.
    MoCAR therefore forecasts directly in code space, without trajectory-space re-tokenization or proposal-and-refinement pipelines.
    }
    \label{fig:code_autoreg}
\end{figure}

Figure~\ref{fig:code_autoreg} illustrates the rollout. Let $C_h$ and $C_f$ denote the numbers of historical and future motion segments. Historical codes are obtained as $\mathbf{z}_{i,c}^{\mathrm{gt}}=\boldsymbol{\mu}_{i,c}$ for $c=1,\ldots,C_h$. Following DONUT's history-conditioning strategy~\citep{knoche2025donut}, these codes form a teacher-forced conditioning prefix. The historical marker in Figure~\ref{fig:code_autoreg} identifies this observed prefix. Teacher forcing applies only to the observed history; future prediction is free-running.

Future forecasting starts from the last historical code. A mode represents one of the $K$ possible future hypotheses. We initialize a mode-specific hidden state:
\begin{equation}
    \mathbf{h}_{i,k,C_h}=W_z\mathbf{z}_{i,C_h}^{\mathrm{gt}}+\mathbf{e}_k,
\end{equation}
Here, $W_z$ projects a motion code into the predictor's hidden dimension. The learned embedding $\mathbf{e}_k$ separates rollout branches and appears as the future mode marker in Figure~\ref{fig:code_autoreg}. At inference, each mode follows a deterministic code rollout conditioned on observed motion and the evolving scene.

At future step $c=C_h+1,\ldots,C_h+C_f$, the predictor updates the hidden state using scene context and predicts the next motion code. The code is decoded into a segment and projected back to hidden space:
\begin{equation}
    \tilde{\mathbf{h}}_{i,k,c}=F_{\Theta}(\mathbf{h}_{i,k,c-1},\mathcal{C}_c), \quad
    \hat{\mathbf{z}}_{i,k,c}=W_o\tilde{\mathbf{h}}_{i,k,c}, \quad
    \hat{\mathcal{S}}_{i,k,c}=D_{\psi}(\hat{\mathbf{z}}_{i,k,c}), \quad
    \mathbf{h}_{i,k,c}=W_z\hat{\mathbf{z}}_{i,k,c}+\mathbf{e}_k .
\end{equation}
Here, $\hat{\cdot}$ denotes predicted quantities. The scene context $\mathcal{C}_c$ includes temporal memory, map elements, neighboring agents, and other modes.

\paragraph{Persistent latent autoregression.}
The predicted motion code $\hat{\mathbf{z}}_{i,k,c}$ directly forms the next hidden state through $W_z$ and the mode embedding. The decoded segment supplies the trajectory and endpoint anchor $\hat{\mathbf{r}}_{i,k,c}$ for the next scene update. Using these codes as autoregressive states lets the predictor model successive local motions together with their changes of reference frame. This persistent code feedback avoids trajectory-space re-tokenization. Each code decodes to a motion segment without an additional refinement stage. Temporal attention separately reuses per-layer caches of projected keys and values from earlier hidden states (Appendix~\ref{sec:appendix_predictor}).

\subsection{Scene-conditioned latent update}
\label{sec:scene_conditioned_update}

Scene context is built around the latest available endpoint anchor. The first future step uses the last historical anchor; later steps use the preceding predicted anchor. The map is encoded into features $\mathbf{G}=\{\mathbf{g}_j\}_{j=1}^{M}$. Each map element is anchored at the midpoint of its entrance boundary, with the entrance direction as its heading. Local interaction graphs are updated around the current anchor during rollout.

Map-to-segment edges connect the current hidden state to nearby map elements. Agent-to-agent edges connect nearby agents within the same mode. Temporal edges connect the current hidden state to the states of historical and previously predicted tokens. Mode-to-mode edges connect the hidden states of different hypotheses for the same agent~\citep{zhou2023query,zhang2024decoupling}. These scene-conditioning components are adapted from prior work. MoCAR's contribution is the coordinate-aware continuous motion-code formulation and its persistent autoregressive feedback.

For an edge from source $u$ to target $v$ with anchor poses $(\mathbf{p}_u,\theta_u)$ and $(\mathbf{p}_v,\theta_v)$~\citep{zhou2023query}, we encode
\begin{equation}
    \mathbf{r}_{u\rightarrow v} =
    \left[
    \|\mathbf{p}_u-\mathbf{p}_v\|_2,\;
    \angle(\mathbf{d}_v,\mathbf{p}_u-\mathbf{p}_v),\;
    \mathrm{wrap}(\theta_u-\theta_v),\;
    \Delta c
    \right],
\end{equation}
Here, $\mathbf{d}_v=(\cos\theta_v,\sin\theta_v)$ is the heading direction, and $\angle(\cdot,\cdot)$ denotes the signed bearing angle. The segment-time offset $\Delta c$ is used only for temporal edges. This encoding describes $u$ in $v$'s local frame. Fourier features embed the relative geometry for attention~\citep{mildenhall2021nerf,tancik2020fourier}.

The predictor applies a stack of context-update blocks to the hidden state. Each block applies temporal, map-to-segment, and agent-to-agent attention in sequence. Mode-to-mode attention is applied once after the final block. Relative geometric encodings modulate the key and value projections in the temporal, map, and agent attention layers.

At the final future prediction step, an MLP produces $\pi_i^{(k)}$ from each mode's updated hidden state. These probabilities score the $K$ complete predicted trajectories. Appendix~\ref{sec:appendix_predictor} provides the detailed computations.

\subsection{Training objectives}
\label{sec:training_objectives}

The motion-code tokenizer and de-tokenizer are trained with a VAE objective
\begin{equation}
\label{eq:vae_loss}    \mathcal{L}_{\mathrm{vae}}=\mathcal{L}_{\mathrm{rec}}+\beta_{\mathrm{kl}}\mathcal{L}_{\mathrm{kl}}, \quad \mathcal{L}_{\mathrm{kl}}=\frac{1}{2}\sum_d(\mu_d^2+\sigma_d^2-\log\sigma_d^2-1),
\end{equation}
Here, $\mathcal{L}_{\mathrm{rec}}$ supervises local segment positions, headings, and velocities. It also includes auxiliary bridge supervision for segment connection (Appendix~\ref{sec:appendix_dec}).

For future trajectory supervision, we use winner-takes-all. The best mode is $k_i^{\star}=\arg\min_k\sum_{t=T_h+1}^{T_h+T_f}\|\hat{\mathbf{p}}_i^{t,k}-\mathbf{p}_i^t\|_2$. This mode receives the position and heading losses
\begin{equation}
    \mathcal{L}_{\mathrm{traj}}=\sum_{i,t}\|\hat{\mathbf{p}}_i^{t,k_i^\star}-\mathbf{p}_i^t\|_1,
    \quad
    \mathcal{L}_{\mathrm{head}}=\sum_{i,t}|\mathrm{wrap}(\hat{\theta}_i^{t,k_i^\star}-\theta_i^t)|,
\end{equation}
where the time sums cover the future horizon.

We also align predicted future codes with encoder means from the corresponding ground-truth segments:
\begin{equation}
    \mathcal{L}_{\mathrm{align}}=\sum_{i,c}\|\hat{\mathbf{z}}_{i,k_i^\star,c}-\boldsymbol{\mu}_{i,c}^{\mathrm{gt}}\|_2^2.
\end{equation}
Here, $c$ indexes future segments. Each target $\boldsymbol{\mu}_{i,c}^{\mathrm{gt}}$ is computed from the ground-truth segment normalized in its own endpoint frame. These encoder means supply training targets; rollout uses predicted codes and anchors.

Mode probabilities are trained with a distance-based soft classification loss
\begin{equation}
    \mathcal{L}_{\mathrm{cls}} = -\sum_i \log \sum_{k=1}^{K}\pi_i^{(k)} \exp\left(-\tau\|\hat{\mathbf{p}}_i^{T_h+T_f,k}-\mathbf{p}_i^{T_h+T_f}\|_2\right).
\end{equation}
Historical segments also receive position and heading supervision. Their weighted loss is $\mathcal{L}_{\mathrm{hist}}$, with coefficients listed in Appendix~\ref{sec:appendix_loss}. The total objective is $\mathcal{L} = \lambda_{\mathrm{vae}}\mathcal{L}_{\mathrm{vae}} + \lambda_{\mathrm{traj}}\mathcal{L}_{\mathrm{traj}} + \lambda_{\mathrm{head}}\mathcal{L}_{\mathrm{head}} + \lambda_{\mathrm{align}}\mathcal{L}_{\mathrm{align}} + \lambda_{\mathrm{cls}}\mathcal{L}_{\mathrm{cls}} + \mathcal{L}_{\mathrm{hist}}$.

\section{Experiments}
\label{sec:experiments}

We first evaluate forecasting accuracy, zero-shot transfer, and performance on turn-heavy scenes. We then examine computational cost and motion-code structure, followed by controlled ablations of the representation and autoregressive feedback.

\subsection{Experimental setup}
\label{sec:setup}

\textbf{Datasets.}
We evaluate on Argoverse~2 (AV2)~\citep{wilson2023argoverse} and Argoverse~1 (AV1)~\citep{chang2019argoverse}. AV2 contains 250k scenarios collected across six cities with 5\,s of history and 6\,s prediction; AV1 has 2\,s of observation and 3\,s prediction. We use AV2 as the primary benchmark; AV1 is used to evaluate cross-dataset transfer and the reusability of the learned motion-code representation across different observation and prediction horizons.

\textbf{Evaluation metrics.}
Following standard protocols~\citep{gu2021densetnt,zhou2022hivt,zhou2023query}, we report minADE$_K$, minFDE$_K$, b-minFDE$_K$, and MR$_K$ with $K=6$ unless otherwise specified. All ablations use the AV2 validation set unless otherwise stated.

\textbf{Implementation details.}
We use $K=6$ modes, hidden dimension 128, 8 attention heads, and 24-D motion codes. Each segment contains $P=16$ points with stride $S=P-1=15$ by default, so the 60-step AV2 future is represented by four codes and the 30-step AV1 future by two. After reconstruction pretraining, the tokenizer and de-tokenizer are jointly optimized with the predictor at a smaller learning rate. During rollout, local map and agent interaction graphs are rebuilt around the current predicted anchor. We train with AdamW and cosine decay, without ensembling or test-time augmentation. Appendix~\ref{sec:appendix_impl} provides the full training configuration, and Appendix~\ref{sec:appendix_av1} details the shared AV1 preprocessing used by all compared methods.

\subsection{Main results}
\paragraph{Comparison with state-of-the-art methods.}
\label{sec:sota}
Table~\ref{tab:sota_av2} compares MoCAR with representative non-ensemble methods on the AV2 test set. MoCAR reaches 1.17 minFDE$_6$, 0.62 minADE$_6$, 0.13 MR$_6$, and 1.83 b-minFDE$_6$. It matches the best MR$_6$, tying or near-tying the remaining multimodal metrics. This performance is achieved with a single-stage decoder-only next-code formulation, without trajectory-space re-tokenization or additional refinement.

\begin{table}[t]
\centering
\caption{Comparison with state-of-the-art non-ensemble methods on the AV2 test set. MoCAR reaches DeMo/DONUT-level performance without trajectory queries or explicit refinement. Best results are in bold, and second-best results are underlined.}
\label{tab:sota_av2}
\scalebox{\uniformtablescale}{
\begin{tabular}{l|cccccc}
\toprule
\textbf{Method}
& \textbf{minFDE$_1$}$\downarrow$
& \textbf{minADE$_1$}$\downarrow$
& \textbf{minFDE$_6$}$\downarrow$
& \textbf{minADE$_6$}$\downarrow$
& \textbf{MR$_6$}$\downarrow$
& \textbf{b-minFDE$_6$}$\downarrow$ \\
\midrule
FRM~\citep{park2023leveraging}
& 5.93 & 2.37 & 1.81 & 0.89 & 0.29 & 2.47 \\
HDGT~\citep{jia2023hdgt}
& 5.37 & 2.08 & 1.60 & 0.84 & 0.21 & 2.24 \\
MTR~\citep{shi2022motion}
& 4.39 & 1.74 & 1.44 & 0.73 & 0.15 & 1.98 \\
HPTR~\citep{zhang2023real}
& 4.61 & 1.84 & 1.43 & 0.73 & 0.19 & 2.03 \\
HeteroGCN~\citep{gao2023dynamic}
& 4.40 & 1.72 & 1.34 & 0.69 & 0.18 & 1.90 \\
ProphNet~\citep{wang2023prophnet}
& 4.74 & 1.80 & 1.33 & 0.68 & 0.18 & 1.88 \\
HiMAP~\citep{xu2026himap}
& 4.61 & 1.81 & 1.33 & 0.68 & 0.17 & 1.96 \\
QCNet~\citep{zhou2023query}
& 4.30 & 1.69 & 1.29 & 0.65 & 0.16 & 1.91 \\
SmartRefine~\citep{zhou2024smartrefine}
& 4.17 & 1.65 & 1.23 & 0.63 & 0.15 & 1.86 \\
DeMo~\citep{zhang2024decoupling}
& \textbf{3.74} & \textbf{1.49} & \underline{1.17} & \textbf{0.61} & \textbf{0.13} & 1.84 \\
DONUT~\citep{knoche2025donut}
& \underline{4.06} & 1.66 & \textbf{1.16} & 0.63 & \underline{0.14} & \textbf{1.79} \\
\midrule
\textbf{MoCAR (Ours)}
& 4.08 & \underline{1.61} & \underline{1.17} & \underline{0.62} & \textbf{0.13} & \underline{1.83} \\
\bottomrule
\end{tabular}
}
\end{table}

\begin{table}[t]
\centering
\caption{Zero-shot transfer to AV1 validation. All methods use AV2-trained checkpoints with identical preprocessing. Full AV1 evaluation appears in Appendix~\ref{sec:appendix_av1}.}
\label{tab:sota_av1}
\scalebox{\uniformtablescale}{
\begin{tabular}{l|cccccc}
\toprule
\textbf{Method (AV2$\rightarrow$AV1, zero-shot)}
& \textbf{minFDE$_1$}$\downarrow$
& \textbf{minADE$_1$}$\downarrow$
& \textbf{minFDE$_6$}$\downarrow$
& \textbf{minADE$_6$}$\downarrow$
& \textbf{MR$_6$}$\downarrow$
& \textbf{b-minFDE$_6$}$\downarrow$ \\
\midrule
DeMo~\citep{zhang2024decoupling} & 5.70 & 2.52 & 2.03 & 1.08 & 0.25 & 2.74 \\
QCNet~\citep{zhou2023query} & 5.60 & 2.51 & \underline{1.55} & \underline{0.95} & \underline{0.19} & 2.27 \\
DONUT~\citep{knoche2025donut} & \textbf{5.05} & \textbf{2.32} & 1.58 & 0.97 & \underline{0.19} & \underline{2.26} \\ \midrule
\textbf{MoCAR (Ours)} & \underline{5.14} & \underline{2.37} & \textbf{1.24} & \textbf{0.72} & \textbf{0.11} & \textbf{1.93} \\
\bottomrule
\end{tabular}
}
\end{table}

\begin{table}[!ht]
\centering
\caption{
Performance on the AV2 validation set stratified by future turning angle.
The turn-heavy subset contains samples whose ground-truth future trajectory has a heading change of at least $45^{\circ}$.
}
\label{tab:turn_subset}
\scalebox{\uniformtablescale}{
\begin{tabular}{llcccc}
\toprule
\textbf{Method}
& \textbf{Subset}
& \textbf{minFDE$_6$}$\downarrow$
& \textbf{minADE$_6$}$\downarrow$
& \textbf{MR$_6$}$\downarrow$
& \textbf{b-minFDE$_6$}$\downarrow$ \\
\midrule
\multirow{3}{*}{QCNet~\citep{zhou2023query}}
& Turn-heavy ($\geq 45^{\circ}$) & 2.37 & 1.17 & 0.36 & 2.98 \\
& Lower-turning ($<45^{\circ}$) & \underline{1.10} & \underline{0.66} & \underline{0.13} & \underline{1.72} \\
& All & 1.25 & \underline{0.72} & 0.16 & 1.87 \\
\midrule
\multirow{3}{*}{DONUT~\citep{knoche2025donut}}
& Turn-heavy ($\geq 45^{\circ}$) & \underline{2.01} & \underline{1.10} & \underline{0.31} & \textbf{2.64} \\
& Lower-turning ($<45^{\circ}$) & \textbf{1.07} & 0.67 & \textbf{0.12} & \textbf{1.70} \\
& All & \underline{1.18} & 0.73 & \underline{0.14} & \textbf{1.81} \\
\midrule
\multirow{3}{*}{MoCAR}
& Turn-heavy ($\geq 45^{\circ}$) & \textbf{1.99} & \textbf{1.07} & \textbf{0.28} & \underline{2.67} \\
& Lower-turning ($<45^{\circ}$) & \textbf{1.07} & \textbf{0.65} & \textbf{0.12} & \underline{1.72} \\
& All & \textbf{1.17} & \textbf{0.70} & \textbf{0.13} & \underline{1.84} \\
\bottomrule
\end{tabular}
}
\end{table}

\paragraph{Cross-dataset zero-shot transfer to AV1.}
If a coordinate-aware motion-code space captures reusable local motion primitives rather than dataset-specific trajectory statistics, it should transfer across datasets. We test this by directly evaluating AV2-trained checkpoints on AV1 without fine-tuning. This setting is challenging because the datasets differ in observation and prediction horizons, available motion states, and agent-label conventions. All methods use the same preprocessing and masking protocol. Velocity and heading are derived from smoothed historical position sequences, and unavailable historical steps are masked out. Metrics are computed over the original 3\,s AV1 future. The shared 10\,Hz sampling rate preserves the 1.5\,s duration of each 16-point segment. Appendix~\ref{sec:appendix_av1} details preprocessing and agent-label mapping.

As shown in Table~\ref{tab:sota_av1}, MoCAR achieves the best results on all four multimodal metrics. It reaches 1.24 minFDE$_6$, compared with 1.55 for QCNet, the strongest external baseline on this metric. This supports our hypothesis that endpoint-normalized continuous motion codes provide a reusable autoregressive vocabulary across different horizons and coordinate conventions. A further ablation shows that training-time positional-noise augmentation benefits zero-shot transfer, consistent with improved robustness to position-derived motion inputs (Appendix~\ref{sec:appendix_av1}).

\paragraph{Performance on turn-heavy scenarios.}
On the AV2 validation set, we evaluate 3,007 cases whose ground-truth future heading change is at least $45^{\circ}$. QCNet and DONUT serve as encoder--decoder and autoregressive baselines. This subset stresses two properties of the forecasting formulation. Large turns require map context and neighboring agents to be updated along the future path, motivating step-wise autoregressive rollout. They also induce large reference-frame changes between consecutive segments, making autoregressive-state consistency important. DONUT updates context during trajectory-space rollout, while MoCAR additionally keeps each generated code as a persistent latent token.

As shown in Table~\ref{tab:turn_subset}, all methods degrade on turn-heavy cases. MoCAR achieves the lowest minFDE$_6$, minADE$_6$, and MR$_6$ among the compared methods and remains comparable on lower-turning cases. These results support coordinate-aware latent autoregression for complex maneuvers. Additional refinement comparisons and qualitative visualizations are provided in Appendices~\ref{sec:appendix_turn_refinement} and~\ref{sec:appendix_qual}.

\begin{table}[t]
\centering
\caption{Parameter count and GPU inference latency on a single Nvidia RTX 4090, measured per scenario with batch size $B=1$. Stride is the number of timesteps advanced per segment and does not apply to QCNet.}
\label{tab:cost}
\scalebox{\uniformtablescale}{
\begin{tabular}{lccc}
\toprule
\textbf{Method} & \textbf{Stride} & \textbf{\#Params} & \textbf{GPU inference (ms)} \\
\midrule
QCNet~\citep{zhou2023query} & -- & 7.7M & 35.6 \\
DONUT~\citep{knoche2025donut} & 10 & 9.0M & 126.7 \\
MoCAR (matched stride) & 10 & 6.3M & 97.9 \\
MoCAR (default) & 15 & 6.3M & 66.8 \\
\bottomrule
\end{tabular}
}
\end{table}

\subsection{Further analysis}

\paragraph{Computational cost.}
\label{sec:appendix_cost}
Table~\ref{tab:cost} reports parameter count and GPU inference latency for representative encoder--decoder and autoregressive baselines, QCNet and DONUT, on a single Nvidia RTX 4090. MoCAR uses per-layer K/V caching and avoids feedback re-encoding, while retaining decoding for trajectories and scene updates. At a matched stride of 10, it takes 97.9\,ms per scenario, a 22.7\% latency reduction relative to DONUT under the same timing protocol. The default stride of 15 uses fewer rollout steps and takes 66.8\,ms. Compared with QCNet, MoCAR incurs extra cost from autoregressive context updates, but retains real-time GPU inference speed ($>10\,\mathrm{fps}$) and supports step-wise map and interaction reasoning under evolving local frames.

\begin{figure}[t]
    \centering
    \includegraphics[width=\linewidth]{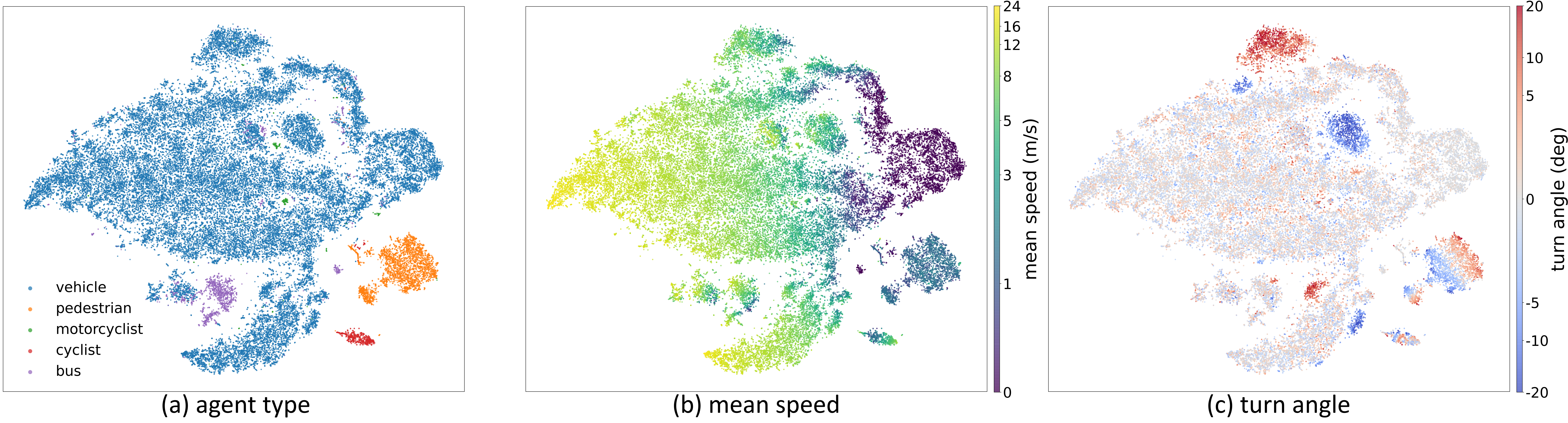}
    \caption{
    \textbf{Learned motion-code space.}
    t-SNE visualization of encoder means from 50,000 AV2 validation segments, colored by agent type, mean speed, and turn angle. The learned space organizes segments by agent category and motion dynamics, suggesting that the latent codes capture fine-grained motion primitives.
    }
    \label{fig:codebook_tsne}
\end{figure}

\paragraph{Structure of the learned motion-code space.}
\label{sec:latent_structure}
To inspect whether the learned codes form a meaningful motion vocabulary, we apply t-SNE~\citep{van2008visualizing} to the encoder means of 50,000 AV2 validation segments (Figure~\ref{fig:codebook_tsne}). The embedding is structured by both agent type and motion dynamics: vehicle segments form the dominant manifold, while pedestrians, cyclists, and other vulnerable road users occupy more peripheral regions. Coloring the same embedding by speed and turn angle reveals smooth local variation rather than fragmented clusters. Pedestrians concentrate in lower-speed regions but exhibit diverse turning directions, whereas vehicles and buses show more regular turn-angle patterns, consistent with stronger constraints from road geometry and traffic rules. Cyclists occupy mixed regions with relatively high speeds and varied turns. These patterns suggest that MoCAR learns a continuous motion vocabulary that preserves fine-grained similarities across agent types, speeds, and curvatures, complementing the quantitative ablations in Section~\ref{sec:ablation}.

\paragraph{Motion-code reconstruction.}
Across motion categories on AV2 validation, mean position reconstruction error over local 1.5\,s segments ranges from 0.02\,m for stationary motion to 0.16\,m for very sharp turns. These results indicate that the codes preserve local motion geometry. Appendix~\ref{sec:appendix_codebook} provides the full breakdown of position and heading errors by motion type.

\subsection{Ablation studies}
\label{sec:ablation}

\paragraph{Motion-code design.}
\label{sec:ablation_code_design}
Table~\ref{tab:ablation_code_design} examines how code capacity and segment granularity affect forecasting accuracy and the number of autoregressive steps. Increasing the code dimension from 16 to 24 improves minFDE$_6$ from 1.23 to 1.17. A further increase to 32 brings no additional gain. Thus, $z=24$ strikes a good balance between representational capacity and predictability. Varying the stride at $z=24$ shows a complementary trade-off. We set $P=S+1$ so adjacent segments always share one boundary point; longer strides shorten the rollout but gradually degrade performance. Strides 10 and 15 give comparable accuracy, and we choose stride 15 for fewer rollout steps and lower latency (Table~\ref{tab:cost}). Both start and endpoint anchoring use local pose normalization. Endpoint anchoring expresses each segment in the latest pose used for the next scene update and improves over start anchoring (Appendix~\ref{sec:appendix_anchor}).

\begin{table}[t]
\centering
\caption{Ablation on motion-code design on the AV2 validation set.}
\label{tab:ablation_code_design}
\scalebox{\uniformtablescale}{
\begin{tabular}{lccccc}
\toprule
& \textbf{Variant}
& \textbf{minFDE$_6$}$\downarrow$
& \textbf{minADE$_6$}$\downarrow$
& \textbf{MR$_6$}$\downarrow$
& \textbf{b-minFDE$_6$}$\downarrow$ \\
\midrule
\multirow{3}{*}{\textit{Code dim}} & $z$ dim = 16 & 1.23 & 0.72 & 0.153 & 1.88 \\
& $z$ dim = 24 & \textbf{1.17} & \textbf{0.70} & \textbf{0.134} & \textbf{1.84} \\
& $z$ dim = 32 & 1.18 & \textbf{0.70} & 0.138 & 1.85 \\
\midrule
\multirow{4}{*}{\textit{Stride}} & stride = 10 & 1.18 & \textbf{0.70} & \textbf{0.133} & \textbf{1.84} \\
& stride = 15 & \textbf{1.17} & \textbf{0.70} & 0.134 & \textbf{1.84} \\
& stride = 20 & 1.18 & \textbf{0.70} & 0.138 & 1.85 \\
& stride = 30 & 1.20 & 0.71 & 0.144 & 1.86 \\
\bottomrule
\end{tabular}
}
\end{table}

\begin{table}[t]
\centering
\caption{Ablation on motion-code representation, training, and feedback on AV2 validation. Best results are in bold.}
\label{tab:ablation_code_space}
\scalebox{\uniformtablescale}{
\begin{tabular}{lcccc}
\toprule
\textbf{Variant}
& \textbf{minFDE$_6$}$\downarrow$
& \textbf{minADE$_6$}$\downarrow$
& \textbf{MR$_6$}$\downarrow$
& \textbf{b-minFDE$_6$}$\downarrow$ \\
\midrule
Raw geometric MLP encoder--decoder & 1.29 & 0.73 & 0.17 & 1.92 \\
Same-architecture GRU (forecasting only) & 1.27 & 0.73 & 0.16 & 1.91 \\
w/o latent-alignment loss $\mathcal{L}_{\mathrm{align}}$ & 1.23 & 0.72 & 0.15 & 1.89 \\
Autoencoder tokenizer ($\beta_{\mathrm{kl}}=0$) & 1.29 & 0.75 & 0.16 & 1.95 \\
Frozen pretrained tokenizer--de-tokenizer & 1.25 & 0.73 & 0.15 & 1.90 \\
MoCAR (re-encoding feedback) & 1.21 & 0.72 & 0.15 & 1.87 \\
\midrule
MoCAR (full) & \textbf{1.17} & \textbf{0.70} & \textbf{0.13} & \textbf{1.84} \\
\bottomrule
\end{tabular}
}
\end{table}

\paragraph{Motion-code learning and feedback.}
\label{sec:ablation_codebook_works}
Table~\ref{tab:ablation_code_space} examines the motion-code representation, its training objectives, and the feedback state used during rollout. We first compare the full model with two forecasting-only controls: a lightweight MLP encoder--decoder over flattened endpoint-normalized step features, and the same GRU tokenizer--de-tokenizer used by MoCAR. Both retain a learned continuous autoregressive interface, but omit reconstruction pretraining, reconstruction loss, KL regularization, and latent alignment. Both controls perform worse than the full model. With the same GRU architecture, the full motion-code training scheme reduces minFDE$_6$ from 1.27 to 1.17. This comparison demonstrates the benefit of the learning scheme beyond the tokenizer architecture.

Removing $\mathcal{L}_{\mathrm{align}}$ degrades forecasting, showing that predicted codes benefit from supervision toward ground-truth encoder codes. Removing the KL regularizer effectively turns the tokenizer into an autoencoder and causes a more pronounced drop, consistent with the benefit of a smooth latent space during rollout. Freezing the pretrained tokenizer--de-tokenizer also hurts, indicating that reconstruction-only pretraining is not fully aligned with forecasting. Together, these results show the importance of latent alignment, weak KL regularization, and joint tokenizer--predictor optimization for latent autoregressive forecasting. A weight comparison in Appendix~\ref{sec:appendix_loss_sensitivity} also favors weak latent alignment over both no alignment and a stronger penalty.

We also compare direct code reuse with re-encoding decoded segments, keeping the predictor, VAE, losses, data, and training schedule fixed. Both variants decode segments for anchor and scene updates; only the state passed to the next prediction changes. Direct reuse improves all four 6\,s metrics, reducing minFDE$_6$ from 1.21 to 1.17. Short-horizon errors are comparable, while the benefit of direct reuse becomes clearer later in the rollout (Appendix~\ref{sec:appendix_feedback}). This controlled comparison supports preserving the predicted motion code as a persistent state over repeated rollout.

\section{Conclusion}
\label{sec:conclusion}

MoCAR formulates trajectory forecasting as next-code prediction over continuous motion codes that jointly represent local trajectory geometry and the induced reference-frame transition. Predicted codes directly condition subsequent predictions, while decoded endpoints update the scene context in a single-stage decoder-only architecture. Experiments show strong AV2 forecasting accuracy and AV2-to-AV1 zero-shot transfer, with improved performance on turn-heavy scenes. MoCAR also achieves lower latency than DONUT at a matched stride. Controlled ablations support joint representation learning and persistent code reuse during rollout. More broadly, MoCAR suggests that endpoint-normalized continuous motion codes can serve as reusable state-transition tokens for autoregressive modeling of physical motion. This abstraction may extend beyond road-agent forecasting, since many robotic, aerial, and embodied systems also require continuous rollout under evolving local coordinate frames.

\paragraph{Limitations.}
Motion-code and frame-estimation errors can accumulate during rollout, particularly for rare or sharp maneuvers. Transfer has been evaluated only from AV2 to AV1 at a shared 10 Hz sampling rate; other rates require temporal resampling or adaptation of the code duration. Mode scores are evaluated for ranking rather than calibrated uncertainty, and their suitability for downstream planning remains untested.

\begin{ack}
This work was supported by the German Research Foundation (DFG) as part of the Research Training Group i.c.sens [GRK2159].
\end{ack}

% \section*{References}
% \bibliographystyle{plainnat}
% \bibliography{bib}

{
\small
\bibliographystyle{unsrt}
\bibliography{bib}

@INPROCEEDINGS{wilson2023argoverse,
  author = {Benjamin Wilson and William Qi and Tanmay Agarwal and John Lambert and Jagjeet Singh and Siddhesh Khandelwal and Bowen Pan and Ratnesh Kumar and Andrew Hartnett and Jhony Kaesemodel Pontes and Deva Ramanan and Peter Carr and James Hays},
  title = {{Argoverse} 2: Next Generation Datasets for Self-Driving Perception and Forecasting},
  booktitle = {Proceedings of the Neural Information Processing Systems Track on Datasets and Benchmarks (NeurIPS Datasets and Benchmarks 2021)},
  year = {2021}
}

@inproceedings{chang2019argoverse,
  title={{Argoverse}: 3D tracking and forecasting with rich maps},
  author={Chang, Ming-Fang and Lambert, John and Sangkloy, Patsorn and Singh, Jagjeet and Bak, Slawomir and Hartnett, Andrew and Wang, De and Carr, Peter and Lucey, Simon and Ramanan, Deva and others},
  booktitle={Proceedings of the IEEE/CVF Conference on Computer Vision and Pattern Recognition},
  pages={8748--8757},
  year={2019}
}

@inproceedings{gao2020vectornet,
  title={{VectorNet}: Encoding {HD} maps and agent dynamics from vectorized representation},
  author={Gao, Jiyang and Sun, Chen and Zhao, Hang and Shen, Yi and Anguelov, Dragomir and Li, Congcong and Schmid, Cordelia},
  booktitle={Proceedings of the IEEE/CVF Conference on Computer Vision and Pattern Recognition},
  pages={11525--11533},
  year={2020}
}

@inproceedings{zhou2023query,
  title={Query-centric trajectory prediction},
  author={Zhou, Zikang and Wang, Jianping and Li, Yung-Hui and Huang, Yu-Kai},
  booktitle={Proceedings of the IEEE/CVF Conference on Computer Vision and Pattern Recognition},
  pages={17863--17873},
  year={2023}
}

@inproceedings{zhou2022hivt,
  title={{HiVT}: Hierarchical vector transformer for multi-agent motion prediction},
  author={Zhou, Zikang and Ye, Luyao and Wang, Jianping and Wu, Kui and Lu, Kejie},
  booktitle={Proceedings of the IEEE/CVF Conference on Computer Vision and Pattern Recognition},
  pages={8823--8833},
  year={2022}
}

@inproceedings{liu2024laformer,
  title={{LAformer}: Trajectory prediction for autonomous driving with lane-aware scene constraints},
  author={Liu, Mengmeng and Cheng, Hao and Chen, Lin and Broszio, Hellward and Li, Jiangtao and Zhao, Runjiang and Sester, Monika and Yang, Michael Ying},
  booktitle={Proceedings of the IEEE/CVF Conference on Computer Vision and Pattern Recognition Workshop},
  pages={2039--2049},
  year={2024}
}

@article{jia2023hdgt,
  title={{HDGT}: Heterogeneous driving graph transformer for multi-agent trajectory prediction via scene encoding},
  author={Jia, Xiaosong and Wu, Penghao and Chen, Li and Liu, Yu and Li, Hongyang and Yan, Junchi},
  journal={IEEE Transactions on Pattern Analysis and Machine Intelligence},
  volume={45},
  number={11},
  pages={13860--13875},
  year={2023},
  publisher={IEEE}
}

@article{gao2023dynamic,
  title={Dynamic scenario representation learning for motion forecasting with heterogeneous graph convolutional recurrent networks},
  author={Gao, Xing and Jia, Xiaogang and Li, Yikang and Xiong, Hongkai},
  journal={IEEE Robotics and Automation Letters},
  volume={8},
  number={5},
  pages={2946--2953},
  year={2023},
  publisher={IEEE}
}

@article{shi2024mtr++,
  title={{MTR++}: Multi-agent motion prediction with symmetric scene modeling and guided intention querying},
  author={Shi, Shaoshuai and Jiang, Li and Dai, Dengxin and Schiele, Bernt},
  journal={IEEE Transactions on Pattern Analysis and Machine Intelligence},
  volume={46},
  number={5},
  pages={3955--3971},
  year={2024},
  publisher={IEEE}
}

@inproceedings{liang2020learning,
  title={Learning lane graph representations for motion forecasting},
  author={Liang, Ming and Yang, Bin and Hu, Rui and Chen, Yun and Liao, Renjie and Feng, Song and Urtasun, Raquel},
  booktitle={European Conference on Computer Vision},
  pages={541--556},
  year={2020},
  organization={Springer}
}

@inproceedings{zeng2021lanercnn,
  title={{LaneRCNN}: Distributed representations for graph-centric motion forecasting},
  author={Zeng, Wenyuan and Liang, Ming and Liao, Renjie and Urtasun, Raquel},
  booktitle={2021 IEEE/RSJ International Conference on Intelligent Robots and Systems (IROS)},
  pages={532--539},
  year={2021},
  organization={IEEE}
}

@article{shi2022motion,
  title={Motion transformer with global intention localization and local movement refinement},
  author={Shi, Shaoshuai and Jiang, Li and Dai, Dengxin and Schiele, Bernt},
  journal={Advances in Neural Information Processing Systems},
  volume={35},
  pages={6531--6543},
  year={2022}
}

@article{zhang2024decoupling,
  title={{DeMo}: Decoupling motion forecasting into directional intentions and dynamic states},
  author={Zhang, Bozhou and Song, Nan and Zhang, Li},
  journal={Advances in Neural Information Processing Systems},
  volume={37},
  pages={106582--106606},
  year={2024}
}

@inproceedings{zhou2024smartrefine,
  title={{SmartRefine}: A scenario-adaptive refinement framework for efficient motion prediction},
  author={Zhou, Yang and Shao, Hao and Wang, Letian and Waslander, Steven L and Li, Hongsheng and Liu, Yu},
  booktitle={Proceedings of the IEEE/CVF Conference on Computer Vision and Pattern Recognition},
  pages={15281--15290},
  year={2024}
}

@inproceedings{choi2023r,
  title={R-Pred: Two-stage motion prediction via tube-query attention-based trajectory refinement},
  author={Choi, Sehwan and Kim, Jungho and Yun, Junyong and Choi, Jun Won},
  booktitle={Proceedings of the IEEE/CVF International Conference on Computer Vision},
  pages={8525--8535},
  year={2023}
}

@inproceedings{wang2023prophnet,
  title={ProphNet: Efficient agent-centric motion forecasting with anchor-informed proposals},
  author={Wang, Xishun and Su, Tong and Da, Fang and Yang, Xiaodong},
  booktitle={Proceedings of the IEEE/CVF Conference on Computer Vision and Pattern Recognition},
  pages={21995--22003},
  year={2023}
}

@article{zhang2023real,
  title={Real-time motion prediction via heterogeneous polyline transformer with relative pose encoding},
  author={Zhang, Zhejun and Liniger, Alexander and Sakaridis, Christos and Yu, Fisher and Gool, Luc Van},
  journal={Advances in Neural Information Processing Systems},
  volume={36},
  pages={57481--57499},
  year={2023}
}

@inproceedings{
park2023leveraging,
title={Leveraging Future Relationship Reasoning for Vehicle Trajectory Prediction},
author={Daehee Park and Hobin Ryu and Yunseo Yang and Jegyeong Cho and Jiwon Kim and Kuk-Jin Yoon},
booktitle={The Eleventh International Conference on Learning Representations },
year={2023},
}

@inproceedings{gu2021densetnt,
  title={{DenseTNT}: End-to-end trajectory prediction from dense goal sets},
  author={Gu, Junru and Sun, Chen and Zhao, Hang},
  booktitle={Proceedings of the IEEE/CVF International Conference on Computer Vision},
  pages={15303--15312},
  year={2021}
}

@inproceedings{zhao2021tnt,
  title={{TNT}: Target-driven trajectory prediction},
  author={Zhao, Hang and Gao, Jiyang and Lan, Tian and Sun, Chen and Sapp, Ben and Varadarajan, Balakrishnan and Shen, Yue and Shen, Yi and Chai, Yuning and Schmid, Cordelia and others},
  booktitle={Conference on Robot Learning},
  pages={895--904},
  year={2021},
  organization={PMLR}
}

@inproceedings{liu2021multimodal,
  title={Multimodal motion prediction with stacked transformers},
  author={Liu, Yicheng and Zhang, Jinghuai and Fang, Liangji and Jiang, Qinhong and Zhou, Bolei},
  booktitle={Proceedings of the IEEE/CVF Conference on Computer Vision and Pattern Recognition},
  pages={7577--7586},
  year={2021}
}

@inproceedings{tang2024hpnet,
  title={{HPNet}: Dynamic trajectory forecasting with historical prediction attention},
  author={Tang, Xiaolong and Kan, Meina and Shan, Shiguang and Ji, Zhilong and Bai, Jinfeng and Chen, Xilin},
  booktitle={Proceedings of the IEEE/CVF Conference on Computer Vision and Pattern Recognition},
  pages={15261--15270},
  year={2024}
}

@inproceedings{aydemir2023adapt,
  title={{ADAPT}: Efficient multi-agent trajectory prediction with adaptation},
  author={Aydemir, G{\"o}rkay and Akan, Adil Kaan and G{\"u}ney, Fatma},
  booktitle={Proceedings of the IEEE/CVF International Conference on Computer Vision},
  pages={8295--8305},
  year={2023}
}

@inproceedings{bhattacharyya2023ssl,
  title={{SSL-Lanes}: Self-supervised learning for motion forecasting in autonomous driving},
  author={Bhattacharyya, Prarthana and Huang, Chengjie and Czarnecki, Krzysztof},
  booktitle={Conference on Robot Learning},
  pages={1793--1805},
  year={2023},
  organization={PMLR}
}

@inproceedings{feng2024unitraj,
  title={UniTraj: A unified framework for scalable vehicle trajectory prediction},
  author={Feng, Lan and Bahari, Mohammadhossein and Amor, Kaouther Messaoud Ben and Zablocki, {\'E}loi and Cord, Matthieu and Alahi, Alexandre},
  booktitle={European Conference on Computer Vision},
  pages={106--123},
  year={2024},
  organization={Springer}
}

@inproceedings{xu2022socialvae,
  title={{SocialVAE}: Human trajectory prediction using timewise latents},
  author={Xu, Pei and Hayet, Jean-Bernard and Karamouzas, Ioannis},
  booktitle={European Conference on Computer Vision},
  pages={511--528},
  year={2022},
  organization={Springer}
}

@inproceedings{xu2024controllable,
  title={Controllable diverse sampling for diffusion based motion behavior forecasting},
  author={Xu, Yiming and Cheng, Hao and Sester, Monika},
  booktitle={2024 IEEE Intelligent Vehicles Symposium (IV)},
  pages={2397--2404},
  year={2024},
  organization={IEEE}
}

@inproceedings{jiang2023motiondiffuser,
  title={{MotionDiffuser}: Controllable multi-agent motion prediction using diffusion},
  author={Jiang, Chiyu and Cornman, Andre and Park, Cheolho and Sapp, Benjamin and Zhou, Yin and Anguelov, Dragomir},
  booktitle={Proceedings of the IEEE/CVF Conference on Computer Vision and Pattern Recognition},
  pages={9644--9653},
  year={2023}
}

@article{liang2023stglow,
  title={{STGlow}: A flow-based generative framework with dual-graphormer for pedestrian trajectory prediction},
  author={Liang, Rongqin and Li, Yuanman and Zhou, Jiantao and Li, Xia},
  journal={IEEE Transactions on Neural Networks and Learning Systems},
  year={2023},
  publisher={IEEE}
}

@inproceedings{yan2025trajflow,
  title={{TrajFlow}: Multi-modal motion prediction via flow matching},
  author={Yan, Qi and Zhang, Brian and Zhang, Yutong and Yang, Daniel and White, Joshua and Chen, Di and Liu, Jiachao and Liu, Langechuan and Zhuang, Binnan and Shi, Shaoshuai and others},
  booktitle={2025 IEEE/RSJ International Conference on Intelligent Robots and Systems (IROS)},
  pages={3923--3928},
  year={2025},
  organization={IEEE}
}

@inproceedings{xiang2024socialcvae,
  title={{SocialCVAE}: Predicting pedestrian trajectory via interaction conditioned latents},
  author={Xiang, Wei and Yin, Haoteng and Wang, He and Jin, Xiaogang},
  booktitle={Proceedings of the AAAI Conference on Artificial Intelligence},
  volume={38},
  number={6},
  pages={6216--6224},
  year={2024}
}

@inproceedings{nayakanti2023wayformer,
  title={Wayformer: Motion forecasting via simple and efficient attention networks},
  author={Nayakanti, Nigamaa and Al-Rfou, Rami and Zhou, Aurick and Goel, Kratarth and Refaat, Khaled S and Sapp, Benjamin},
  booktitle={2023 IEEE International Conference on Robotics and Automation (ICRA)},
  pages={2980--2987},
  year={2023},
  organization={IEEE}
}

@article{vaswani2017attention,
  title={Attention is all you need},
  author={Vaswani, Ashish and Shazeer, Noam and Parmar, Niki and Uszkoreit, Jakob and Jones, Llion and Gomez, Aidan N and Kaiser, {\L}ukasz and Polosukhin, Illia},
  journal={Advances in Neural Information Processing Systems},
  volume={30},
  year={2017}
}

@article{tancik2020fourier,
  title={Fourier features let networks learn high frequency functions in low dimensional domains},
  author={Tancik, Matthew and Srinivasan, Pratul and Mildenhall, Ben and Fridovich-Keil, Sara and Raghavan, Nithin and Singhal, Utkarsh and Ramamoorthi, Ravi and Barron, Jonathan and Ng, Ren},
  journal={Advances in Neural Information Processing Systems},
  volume={33},
  pages={7537--7547},
  year={2020}
}

@article{mildenhall2021nerf,
  title={NeRF: Representing scenes as neural radiance fields for view synthesis},
  author={Mildenhall, Ben and Srinivasan, Pratul P and Tancik, Matthew and Barron, Jonathan T and Ramamoorthi, Ravi and Ng, Ren},
  journal={Communications of the ACM},
  volume={65},
  number={1},
  pages={99--106},
  year={2021},
  publisher={ACM New York, NY, USA}
}

@inproceedings{knoche2025donut,
  title={DONUT: A Decoder-Only Model for Trajectory Prediction},
  author={Knoche, Markus and de Geus, Daan and Leibe, Bastian},
  booktitle={Proceedings of the IEEE/CVF International Conference on Computer Vision},
  pages={28903--28912},
  year={2025}
}

@inproceedings{
philion2023trajeglish,
title={Trajeglish: Traffic Modeling as Next-Token Prediction},
author={Jonah Philion and Xue Bin Peng and Sanja Fidler},
booktitle={The Twelfth International Conference on Learning Representations},
year={2024},
}

@article{kingma2013auto,
  title={Auto-encoding variational bayes},
  author={Kingma, Diederik P and Welling, Max},
  journal={arXiv preprint arXiv:1312.6114},
  year={2013}
}

@article{van2017neural,
  title={Neural discrete representation learning},
  author={van den Oord, Aaron and Vinyals, Oriol and Kavukcuoglu, Koray},
  journal={Advances in neural information processing systems},
  volume={30},
  year={2017}
}

@article{xu2026himap,
  title={{HiMAP}: History-aware Map-occupancy Prediction with Fallback},
  author={Xu, Yiming and Yang, Yi and Cheng, Hao and Sester, Monika},
  journal={arXiv preprint arXiv:2602.17231},
  year={2026}
}

@inproceedings{yao2024improving,
  title={Improving out-of-distribution generalization of trajectory prediction for autonomous driving via polynomial representations},
  author={Yao, Yue and Yan, Shengchao and Goehring, Daniel and Burgard, Wolfram and Reichardt, Joerg},
  booktitle={2024 IEEE/RSJ International Conference on Intelligent Robots and Systems (IROS)},
  pages={488--495},
  year={2024},
  organization={IEEE}
}

@article{van2008visualizing,
  title={Visualizing data using t-{SNE}.},
  author={Van der Maaten, Laurens and Hinton, Geoffrey},
  journal={Journal of machine learning research},
  volume={9},
  number={11},
  year={2008}
}

@inproceedings{seff2023motionlm,
  title={{MotionLM}: Multi-agent motion forecasting as language modeling},
  author={Seff, Ari and Cera, Brian and Chen, Dian and Ng, Mason and Zhou, Aurick and Nayakanti, Nigamaa and Refaat, Khaled S and Al-Rfou, Rami and Sapp, Benjamin},
  booktitle={Proceedings of the IEEE/CVF International Conference on Computer Vision},
  pages={8579--8590},
  year={2023}
}

@inproceedings{wu2024smart,
title={{SMART}: Scalable Multi-agent Real-time Motion Generation via Next-token Prediction},
author={Wei Wu and Xiaoxin Feng and Ziyan Gao and Yuheng KAN},
booktitle={The Thirty-eighth Annual Conference on Neural Information Processing Systems},
year={2024},
}

@article{zhou2024behaviorgpt,
  title={Behaviorgpt: Smart agent simulation for autonomous driving with next-patch prediction},
  author={Zhou, Zikang and Hu, Haibo and Chen, Xinhong and Wang, Jianping and Guan, Nan and Wu, Kui and Li, Yung-Hui and Huang, Yu-Kai and Xue, Chun J},
  journal={Advances in Neural Information Processing Systems},
  volume={37},
  pages={79597--79617},
  year={2024}
}
}

% References follow the acknowledgments in the camera-ready paper. Use unnumbered first-level heading for
% the references. Any choice of citation style is acceptable as long as you are
% consistent. It is permissible to reduce the font size to \verb+small+ (9 point)
% when listing the references.
% Note that the Reference section does not count towards the page limit.
% \medskip

% {
% \small

% [1] Alexander, J.A.\ \& Mozer, M.C.\ (1995) Template-based algorithms for
% connectionist rule extraction. In G.\ Tesauro, D.S.\ Touretzky and T.K.\ Leen
% (eds.), {\it Advances in Neural Information Processing Systems 7},
% pp.\ 609--616. Cambridge, MA: MIT Press.

% [2] Bower, J.M.\ \& Beeman, D.\ (1995) {\it The Book of GENESIS: Exploring
%   Realistic Neural Models with the GEneral NEural SImulation System.}  New York:
% TELOS/Springer--Verlag.

% [3] Hasselmo, M.E., Schnell, E.\ \& Barkai, E.\ (1995) Dynamics of learning and
% recall at excitatory recurrent synapses and cholinergic modulation in rat
% hippocampal region CA3. {\it Journal of Neuroscience} {\bf 15}(7):5249-5262.
% }

%%%%%%%%%%%%%%%%%%%%%%%%%%%%%%%%%%%%%%%%%%%%%%%%%%%%%%%%%%%%

\appendix

\section{Implementation details}
\label{sec:appendix_impl}

\subsection{Motion segments and step features}
\label{sec:appendix_segments}

We use $P=16$ points per segment with a stride $S=15$, so adjacent segments share one boundary point. With AV2's $T_h=50$ historical and $T_f=60$ future steps, this defines four historical segment slots and four future anchors covering the prediction horizon. The last historical slot corresponds to the current state. The number of valid historical motion codes depends on the available observations.
% Segments whose anchor falls outside the observation window or that contain too few valid points are masked out.
When a segment contains fewer than $P$ valid points near sequence boundaries or due to missing observations, invalid positions are padded and excluded by the step mask. Segments with at least three valid points are still encoded by the tokenizer. Its GRU reads only valid steps through the step mask, and reconstruction losses are computed only on valid points. Segments with fewer than three valid points are masked out.

For each segment, we extract per-step transition features in the endpoint-anchor frame: the local position, local heading, step displacement and its direction, heading change, speed magnitude, and velocity direction relative to the anchor heading. These 8-D features provide a redundant but compact description of step-level kinematics that the tokenizer compresses into a single 24-D code.

We additionally supervise a \emph{bridge target}: the local position and heading of the segment's first point in the endpoint-anchor frame. By construction, this equals the previous segment's endpoint pose expressed in the current frame, and is used by an auxiliary head of the de-tokenizer (Section~\ref{sec:appendix_dec}) to enforce inter-segment continuity.

\subsection{VAE encoder (tokenizer)}
\label{sec:appendix_enc}

The tokenizer takes the per-step features of a segment together with an agent-type embedding and produces a Gaussian latent $\mathcal{N}(\boldsymbol{\mu},\mathrm{diag}(\boldsymbol{\sigma}^2))$ with dimension 24. Each 8-D step feature is lifted to a hidden dimension $D=128$ via a Fourier embedding~\cite{mildenhall2021nerf,tancik2020fourier} (64 frequency bands) and a per-dimension MLP. The resulting sequence is processed by a single-layer GRU. For variable-length segments, the step mask determines the valid sequence length; padded transitions are ignored by packed GRU encoding. The final hidden state, combined with the agent-type embedding, is mapped to $\boldsymbol{\mu}$ and $\boldsymbol{\sigma}$ by two linear heads.

\subsection{VAE decoder (de-tokenizer)}
\label{sec:appendix_dec}

The de-tokenizer takes a 24-D code $z$ and reconstructs the segment in the endpoint-anchor frame. It maintains $P$ learned timestep embeddings. Each embedding is concatenated with $z$ and lifted to dimension $D$ by a backbone MLP. A GRU processes the resulting length-$P$ feature sequence. Three heads act on this sequence to predict backward displacements $\Delta\mathbf{p}^{\mathrm{back}}_j\in\mathbb{R}^2$, velocity, and heading. An auxiliary \emph{bridge} head maps $z$ directly to the segment's first-point local pose.

The local relative positions are reconstructed by reverse cumulative summation of the backward displacements,
\begin{equation}
    \tilde{\mathbf{p}}^{j} = -\sum_{j'=j+1}^{P-1} \Delta\mathbf{p}^{\mathrm{back}}_{j'},
\end{equation}
which guarantees $\tilde{\mathbf{p}}^{P-1}=\mathbf{0}$ exactly. Errors thus accumulate \emph{toward the past} rather than away from a noisy first point, keeping the endpoint anchor as a precise integration starting point. The bridge head is functionally redundant with $\tilde{\mathbf{p}}^{0}$ and is used only as an auxiliary training signal to encourage segment-to-segment continuity. It is not used during autoregressive rollout; endpoint-anchor updates are computed from the decoded displacement and heading sequence.

\paragraph{Bridge supervision.}
For a valid pair of adjacent segments indexed by $i$, let $(\mathbf{b}_i,\phi_i)$ be the preceding endpoint pose expressed in the current segment's endpoint frame, and let $(\hat{\mathbf{b}}_i,\hat{\phi}_i)$ be the auxiliary head's prediction. The bridge loss is
\begin{equation}
\begin{aligned}
\mathcal{L}_{\mathrm{bridge}}
&=\mathcal{L}_{\mathrm{bridge,pos}}+\mathcal{L}_{\mathrm{bridge,heading}},\\
\mathcal{L}_{\mathrm{bridge,pos}}
&=\frac{1}{|\mathcal{V}|}\sum_{i\in\mathcal{V}}w_i\|\hat{\mathbf{b}}_i-\mathbf{b}_i\|_2^2,\\
\mathcal{L}_{\mathrm{bridge,heading}}
&=\frac{1}{|\mathcal{V}|}\sum_{i\in\mathcal{V}}w_i\operatorname{wrap}(\hat{\phi}_i-\phi_i)^2,
\end{aligned}
\label{eq:bridge_loss}
\end{equation}
where $\mathcal{V}$ contains valid adjacent-segment pairs and $w_i$ is the same turn-dependent sample weight used for segment reconstruction. The wrapping operation maps the angular difference to $[-\pi,\pi)$. The weight is computed from the target bridge heading:
\begin{equation}
    w_i=1+\alpha\min\!\left(\frac{\max(|\phi_i|-\theta_0,\,0)}{\theta_s},\,2\right),
    \qquad \theta_0=\frac{\pi}{20}, \quad \theta_s=0.35.
    \label{eq:turn_weight}
\end{equation}
All angles are in radians, so $\theta_0$ corresponds to $9^{\circ}$. The angular score is clipped to $[0,2]$. We set $\alpha=1.0$, giving $w_i\in[1,3]$. Bridge supervision is included in $\mathcal{L}_{\mathrm{rec}}$ and encourages continuity between adjacent reconstructed segments; it adds no operation to autoregressive inference.

\subsection{Decoder-only predictor}
\label{sec:appendix_predictor}

The predictor stacks $L=5$ context-update blocks. Each block consists of three sparse-graph attention layers in order: temporal memory, map-to-segment, and agent-to-agent. After the final context-update block, a mode-to-mode attention layer is applied once to exchange information among the $K$ modes of the same agent. Each temporal, map-to-segment, and agent-to-agent layer is a Pre-LN transformer block with $H=8$ heads, head dimension 16, an FFN ratio of $4\times$, and dropout 0.1. Relative geometric features modulate keys and values in these three layers via a Fourier embedding~\citep{mildenhall2021nerf,tancik2020fourier}. The mode-to-mode layer uses multi-head attention over mode states without relative geometric edge attributes. Map-to-segment and agent-to-agent edges are built by radius search around the current anchor with $r=50$ m. Temporal edges are causal and connect each token to all earlier tokens of the same agent.

\paragraph{Temporal KV cache.}
For each temporal attention layer, the predictor caches the projected keys and values of earlier token states. Subsequent rollout steps reuse these projections, while the relative geometric terms are computed from the source and current target anchors. Specifically, the cached terms are $\mathbf{W}^{m}_{K,\mathrm{temp}}\mathrm{LN}(\mathbf{s}_u)$ and $\mathbf{W}^{m}_{V,\mathrm{temp}}\mathrm{LN}(\mathbf{s}_u)$ in the attention equations below. The predicted motion code separately determines the next autoregressive input through $W_z$ and the mode embedding (Section~\ref{sec:code_autoreg}); persistent code feedback and temporal caching therefore serve distinct roles.

\paragraph{Mode initialization.}
After the last historical token, the predictor branches into $K=6$ modes by adding a learned mode embedding to the projected historical code. The output projection $W_o$ that maps hidden states back to $z$ is initialized to zero so that early predicted codes start near the latent-space mean, which we found important for stable rollout.

\paragraph{Attention layer details.}
The temporal, map-to-segment, and agent-to-agent layers use the same sparse relative-attention form, with different source sets. 
For a target token $v$ with hidden state $\mathbf{h}_v$, let $\mathcal{N}_{a}(v)$ denote its sparse source neighborhood under attention type $a \in \{\mathrm{temp}, \mathrm{map}, \mathrm{agent}\}$. 
A source element $u \in \mathcal{N}_{a}(v)$ has feature $\mathbf{s}_u$. For temporal attention, this is a previous token's hidden state. For map-to-segment and agent-to-agent attention, it is a map-element feature and a neighboring-agent hidden state, respectively.
Each edge $(u,v)$ carries a relative geometric encoding $\mathbf{r}^{a}_{u\rightarrow v}$, defined from the source and target anchor poses as in Section~\ref{sec:scene_conditioned_update}. 
This encoding is embedded by Fourier features and a relation-specific MLP:
\begin{equation}
    \boldsymbol{\rho}^{a}_{u\rightarrow v}
    =
    \mathrm{MLP}_{a}
    \left(
    \mathrm{Fourier}(\mathbf{r}^{a}_{u\rightarrow v})
    \right).
\end{equation}

We use $\mathrm{LN}(\cdot)$ to denote layer normalization. 
For attention type $a$ and head $m$, 
$\mathbf{W}^{m}_{Q,a}$, $\mathbf{W}^{m}_{K,a}$, and $\mathbf{W}^{m}_{V,a}$ are learnable query, key, and value projection matrices, respectively. 
The matrices $\mathbf{U}^{m}_{K,a}$ and $\mathbf{U}^{m}_{V,a}$ project the relative edge embedding into the key and value spaces. 
Thus, the query is computed from the target token, while the key and value are computed from the source feature and modulated by the relative geometry:
\begin{equation}
\begin{aligned}
    \mathbf{q}_{v}^{m,a} 
    &= 
    \mathbf{W}^{m}_{Q,a}\,\mathrm{LN}(\mathbf{h}_v), \\
    \mathbf{k}_{u\rightarrow v}^{m,a} 
    &= 
    \mathbf{W}^{m}_{K,a}\,\mathrm{LN}(\mathbf{s}_u)
    +
    \mathbf{U}^{m}_{K,a}\boldsymbol{\rho}^{a}_{u\rightarrow v}, \\
    \mathbf{v}_{u\rightarrow v}^{m,a} 
    &= 
    \mathbf{W}^{m}_{V,a}\,\mathrm{LN}(\mathbf{s}_u)
    +
    \mathbf{U}^{m}_{V,a}\boldsymbol{\rho}^{a}_{u\rightarrow v}.
\end{aligned}
\end{equation}
Here, $\mathbf{q}_{v}^{m,a}$ is the query of target $v$, while $\mathbf{k}_{u\rightarrow v}^{m,a}$ and $\mathbf{v}_{u\rightarrow v}^{m,a}$ are the edge-specific key and value passed from source $u$ to target $v$.

The attention weights are normalized over the sparse source neighborhood of $v$, and the head output is the weighted sum of the modulated values:
\begin{equation}
\begin{aligned}
    \alpha_{u\rightarrow v}^{m,a}
    &=
    \mathrm{softmax}_{u\in\mathcal{N}_{a}(v)}
    \left(
    \frac{
    (\mathbf{q}_{v}^{m,a})^\top \mathbf{k}_{u\rightarrow v}^{m,a}
    }{\sqrt{d_h}}
    \right), \\
    \mathbf{o}_{v}^{m,a}
    &=
    \sum_{u\in\mathcal{N}_{a}(v)}
    \alpha_{u\rightarrow v}^{m,a}
    \mathbf{v}_{u\rightarrow v}^{m,a}.
\end{aligned}
\end{equation}
Here, $\mathbf{o}_{v}^{m,a}$ denotes the output of attention head $m$ for target token $v$ under attention type $a$, obtained as the weighted sum of edge-specific value vectors from its sparse source neighborhood, and $d_h$ is the dimension of each attention head.

The outputs of all $H$ heads are concatenated, projected by the learnable output matrix $\mathbf{W}_{O,a}$, and added back to the target token through a residual connection. 
The result is then passed through a feed-forward network $\mathrm{FFN}_{a}$ with another residual connection:
\begin{equation}
\begin{aligned}
    \bar{\mathbf{h}}_{v}
    &=
    \mathbf{h}_{v}
    +
    \mathbf{W}_{O,a}
    [\mathbf{o}_{v}^{1,a};\ldots;\mathbf{o}_{v}^{H,a}], \\
    \mathbf{h}_{v}^{+}
    &=
    \bar{\mathbf{h}}_{v}
    +
    \mathrm{FFN}_{a}(\mathrm{LN}(\bar{\mathbf{h}}_{v})).
\end{aligned}
\end{equation}
Thus, relative geometry does not form an additional token; instead, each sparse edge contributes an edge embedding that modulates the key and value projections. 
The final mode-to-mode layer uses the same multi-head attention structure over mode states of the same agent, but without relative geometric modulation.

% \paragraph{Auxiliary deep supervision.}
% The last two attention modules each carry an auxiliary head $W_o^{(\ell)}$ that produces an intermediate predicted code, fed through the de-tokenizer and supervised by all reconstruction, trajectory, and alignment losses with a geometric layer-decay weight ($\lambda_{\mathrm{last}}=0.5$, $\rho=0.5$).

\subsection{Loss weights}
\label{sec:appendix_loss}

\paragraph{Loss weights.}
Concrete weights are listed in Table~\ref{tab:loss_weights}. Future-trajectory L1 regression receives the largest weight, while latent alignment has a much smaller coefficient. Despite this small weight, removing alignment increases forecasting error (Table~\ref{tab:ablation_code_space}), supporting its role as auxiliary supervision for the predicted motion codes.

\begin{table}[h]
\centering
\caption{Loss weight settings.}
\label{tab:loss_weights}
\scalebox{\uniformtablescale}{
\begin{tabular}{lc}
\toprule
\textbf{Term} & \textbf{Value} \\
\midrule
VAE objective $\lambda_{\mathrm{vae}}$ & 0.1 \\
KL coefficient $\beta_{\mathrm{kl}}$ & $10^{-4}$ \\
Future-trajectory L1 (winner) $\lambda_{\mathrm{traj}}$ & 5.0 \\
Historical-trajectory L1 $\lambda_{\mathrm{traj\,hist}}$ & 1.0 \\
Heading-trajectory L1 (hist / fut) & 1.0 / 1.0 \\
Latent alignment $\lambda_{\mathrm{align}}$ & 0.01 \\
Soft classification $\lambda_{\mathrm{cls}}$ & 1.0 \\
Soft-cls temperature $\tau$ & 1.0 \\
\bottomrule
\end{tabular}
}
\end{table}

\paragraph{Representation-loss weights.}
\label{sec:appendix_loss_sensitivity}
Table~\ref{tab:loss_sensitivity} complements the removal ablations in Table~\ref{tab:ablation_code_space} with a stronger latent-alignment weight. The architecture, reconstruction objective, pretraining, and joint optimization are retained. Weak KL regularization improves all four forecasting metrics relative to the autoencoder tokenizer. For latent alignment, the default weight of 0.01 outperforms both no alignment and the stronger weight of 1.0. These results support weak latent supervision as an auxiliary training signal. All other loss weights remain at their default values.

\begin{table}[h]
\centering
\caption{Representation-loss weight comparisons on the AV2 validation set. Each group changes one coefficient while keeping the others at their defaults. The zero-weight and default results are repeated from Table~\ref{tab:ablation_code_space}.}
\label{tab:loss_sensitivity}
\scalebox{\uniformtablescale}{
\begin{tabular}{llcccc}
\toprule
\textbf{Term} & \textbf{Weight}
& \textbf{minFDE$_6$}$\downarrow$ & \textbf{minADE$_6$}$\downarrow$
& \textbf{MR$_6$}$\downarrow$ & \textbf{b-minFDE$_6$}$\downarrow$ \\
\midrule
\multirow{2}{*}{KL $\beta_{\mathrm{kl}}$}
& 0 & 1.29 & 0.75 & 0.16 & 1.95 \\
& $10^{-4}$ (default) & 1.17 & 0.70 & 0.13 & 1.84 \\
\midrule
\multirow{3}{*}{Alignment $\lambda_{\mathrm{align}}$}
& 0 & 1.23 & 0.72 & 0.15 & 1.89 \\
& 0.01 (default) & 1.17 & 0.70 & 0.13 & 1.84 \\
& 1.0 & 1.19 & 0.71 & 0.14 & 1.86 \\
\bottomrule
\end{tabular}
}
\end{table}

\subsection{Training schedule}
\label{sec:appendix_train}

\paragraph{Tokenizer pretraining.}
The tokenizer and de-tokenizer are first pretrained for segment reconstruction before being loaded into the full MoCAR model. This stage optimizes only the VAE objective in Eq.~\eqref{eq:vae_loss}, including segment reconstruction, bridge reconstruction, and the weak KL regularizer. We use AdamW with weight decay $10^{-4}$, learning rate $10^{-4}$, batch size $256$, and cosine annealing for $100$ epochs. The KL coefficient is set to $\beta_{\mathrm{kl}}=10^{-4}$. After pretraining, the encoder mean $\boldsymbol{\mu}$ is used as the deterministic motion code for ground-truth segments.

\paragraph{Joint forecasting training.}
The pretrained tokenizer--de-tokenizer is then loaded into the full MoCAR model and jointly optimized with the predictor and map encoder. We use a total batch size of 16 across both GPUs. Training uses AdamW with weight decay $10^{-4}$, an initial learning rate of $2\times10^{-4}$, and cosine annealing to 0 over 64 epochs. The tokenizer--de-tokenizer parameters are placed in a separate parameter group whose learning rate is scaled by $0.1\times$, so that the motion-code space can adapt to forecasting while preserving its reconstruction structure. The full objective combines trajectory regression, heading supervision, soft mode classification, latent alignment, and the VAE loss as described in Section~\ref{sec:training_objectives}.

\paragraph{Positional-noise augmentation.}
During joint training we add Gaussian noise with $\sigma=0.2$ to the positional channels of the input step features for 20\% of agents in each mini-batch. This augmentation is applied only during training and is disabled during validation and testing. Its measured benefit appears in AV2-to-AV1 zero-shot transfer, while AV2 validation metrics are unchanged at the reported precision when it is removed (Section~\ref{sec:appendix_noise}). We do not use ensembling or test-time augmentation.

\paragraph{Hardware.}
Training was conducted on $2\times$ Nvidia RTX Pro 6000 GPUs for approximately 40 hours per full forecasting run.

\paragraph{Inference timing.}
All methods in Table~\ref{tab:cost} are timed on a single NVIDIA RTX 4090 with batch size 1. The matched-stride and default-stride measurements use the same timing protocol. After warming up on 10 scenarios, we measure GPU execution time over the test set. Data loading and preprocessing are excluded. The reported latency is the total measured GPU time divided by the number of test scenarios, in milliseconds per scenario.

\section{Additional ablations}
\label{sec:appendix_ablation}

\subsection{Predictor depth sweep}
\label{sec:appendix_depth}

Table~\ref{tab:appendix_depth} reports an ablation on the number of context-update blocks in the decoder-only predictor. Each block consists of temporal, map-to-segment, and agent-to-agent attention, with a single mode-to-mode attention layer applied after the final block. Increasing depth from three to five consistently improves all reported metrics. The gains largely saturate at five blocks, with little further improvement from a sixth. We therefore use five blocks as the default.

\begin{table}[h]
\centering
\caption{Predictor depth sweep on the AV2 validation set.}
\label{tab:appendix_depth}
\scalebox{\uniformtablescale}{
\begin{tabular}{lcccc}
\toprule
\textbf{Variant}
& \textbf{minFDE$_6$}$\downarrow$
& \textbf{minADE$_6$}$\downarrow$
& \textbf{MR$_6$}$\downarrow$
& \textbf{b-minFDE$_6$}$\downarrow$ \\
\midrule
3 context-update blocks & 1.27 & 0.74 & 0.156 & 1.91 \\
4 context-update blocks & 1.22 & 0.72 & 0.142 & 1.89 \\
5 context-update blocks (MoCAR) & \textbf{1.17} & \textbf{0.70} & 0.134 & \textbf{1.84} \\
6 context-update blocks  & \textbf{1.17} & \textbf{0.70} & \textbf{0.133} & \textbf{1.84} \\
\bottomrule
\end{tabular}
}
\end{table}

\subsection{Choice of coordinate anchor}
\label{sec:appendix_anchor}

Table~\ref{tab:anchor_ablation} compares the choice of local reference pose while retaining local pose normalization and the motion-code formulation. Endpoint anchoring achieves lower minFDE$_6$ than start anchoring on both AV2 validation and AV1 zero-shot evaluation. It also expresses each code in the latest predicted pose used for the next scene update. A start anchor instead refers to the pose 1.5\,s earlier at the default stride.

\begin{table}[h]
\centering
\caption{Choice of segment anchor with local coordinate normalization. Both variants are trained on AV2 and evaluated on AV2 validation and AV1 validation without fine-tuning.}
\label{tab:anchor_ablation}
\scalebox{\uniformtablescale}{
\begin{tabular}{lccccc}
\toprule
& \multicolumn{4}{c}{\textbf{AV2 validation}} & \textbf{AV1 zero-shot} \\
\cmidrule(lr){2-5}\cmidrule(lr){6-6}
\textbf{Anchor} & \textbf{minFDE$_6$}$\downarrow$ & \textbf{minADE$_6$}$\downarrow$
& \textbf{MR$_6$}$\downarrow$ & \textbf{b-minFDE$_6$}$\downarrow$ & \textbf{minFDE$_6$}$\downarrow$ \\
\midrule
Segment start & 1.19 & 0.71 & 0.14 & 1.85 & 1.25 \\
Segment endpoint & 1.17 & 0.70 & 0.13 & 1.84 & 1.24 \\
\bottomrule
\end{tabular}
}
\end{table}

\subsection{Autoregressive feedback across prediction horizons}
\label{sec:appendix_feedback}

Table~\ref{tab:feedback_horizons} complements the full-horizon metrics in Table~\ref{tab:ablation_code_space}. The variants use the same predictor, VAE, losses, data, and training schedule; only the state passed to the next prediction step changes. Both decode segments to update the anchor and scene context. Re-encoding gives slightly lower error at 1.5\,s, and the variants tie at 3\,s. Persistent code reuse has lower error at 4.5\,s and 6\,s, showing the benefit of persistent code reuse over longer rollouts.

\begin{table}[h]
\centering
\caption{Autoregressive feedback ablation on AV2 validation, reporting minFDE$_6$ by prediction horizon.}
\label{tab:feedback_horizons}
\scalebox{\uniformtablescale}{
\begin{tabular}{lcccc}
\toprule
\textbf{Feedback state} & \textbf{1.5\,s} & \textbf{3.0\,s} & \textbf{4.5\,s} & \textbf{6.0\,s} \\
\midrule
Trajectory: decode and re-encode & 0.19 & 0.44 & 0.80 & 1.21 \\
Persistent motion code (MoCAR) & 0.20 & 0.44 & 0.78 & 1.17 \\
\bottomrule
\end{tabular}
}
\end{table}

\subsection{Discrete motion-code control}
\label{sec:appendix_vq}

As a supplementary representation control, we jointly train a VQ-VAE tokenizer--de-tokenizer with the predictor. The latent-alignment target is the codebook entry selected by the future encoder, while the predictor remains free to select another entry. We use 16,384 24-D codes with commitment weight 0.25 and codebook decay 0.99. Initialization uses 20 k-means iterations (\texttt{vq\_init\_ratio}=128); code maintenance uses \texttt{dead\_threshold}=1.0 and \texttt{reset\_interval}=200.

This configuration obtains 1.35 minFDE$_6$, 0.75 minADE$_6$, 0.18 MR$_6$, and 2.02 b-minFDE$_6$ on AV2 validation, with 99.8\% code utilization. The continuous VAE achieves lower forecasting error, supporting its use as the motion-code representation for MoCAR.

\section{Mode utilization}
\label{sec:appendix_perhorizon}

To examine how frequently each mode is selected as the best match to the ground truth, Table~\ref{tab:mode_util} reports the empirical winner-mode distribution and its entropy $H=-\sum_k p_k\log p_k$ over the AV2 validation set. A uniform distribution gives $H=\ln 6\approx1.79$; assigning every sample's winner to the same mode gives $H=0$. The observed entropy of 1.774 is close to the maximum. All six modes are selected across the validation set, and no single mode dominates the winner distribution.

\begin{table}[t]
\centering
\caption{Winner-mode distribution and entropy on the AV2 validation set.}
\label{tab:mode_util}
\scalebox{\uniformtablescale}{
\begin{tabular}{l|cccccc|c}
\toprule
& \textbf{$k\!=\!1$} & \textbf{2} & \textbf{3} & \textbf{4} & \textbf{5} & \textbf{6} & \textbf{Entropy $H$} \\
\midrule
MoCAR & 0.177 & 0.134 & 0.119 & 0.172 & 0.200 & 0.199 & 1.774 \\
\bottomrule
\end{tabular}
}
\end{table}

\section{Motion-code space analysis}
\label{sec:appendix_codebook}

We evaluate the reconstruction quality of the learned VAE motion codes on 681,968 AV2 validation segments across different motion types. The categories in Table~\ref{tab:recon_breakdown} are defined at the segment level: stationary segments are grouped by mean speed, while non-stationary segments are grouped by heading change over a local $P=16$ segment, corresponding to 1.5\,s on AV2. These local categories differ from the turn-heavy forecasting subset, which uses heading change over the full 6\,s future.

As shown in Table~\ref{tab:recon_breakdown}, stationary and straight segments can be reconstructed with very small errors, while the reconstruction error gradually increases as the local turning angle grows. This trend is expected since turning segments contain stronger nonlinear motion and larger heading variation. Nevertheless, mean position error remains low even for sharp and very sharp local turns, at or below 0.16\,m in every category. These results suggest that the continuous VAE latent space can represent diverse motion primitives from lane-following motion to challenging turning behavior. Heading error reaches 0.19\,rad for very sharp turns. The segment counts also highlight the long-tailed nature of local motion: only 860 segments turn by at least $45^{\circ}$ within 1.5\,s.

\begin{table}[t]
\centering
\caption{
VAE reconstruction error on AV2 validation segments, grouped by motion type.
Stationary segments are grouped by mean speed, while non-stationary segments are grouped by segment-level heading change.
Position error is measured by the mean L2 distance over segment points, and heading error is measured by the mean absolute wrapped angle error.
Turn categories are defined over local $P=16$ segments, about 1.5s on AV2; thus $|\Delta\theta|\geq22.5^{\circ}$ already corresponds to a sharp local turn.
}
\label{tab:recon_breakdown}
\scalebox{\uniformtablescale}{
\begin{tabular}{lccc}
\toprule
\textbf{Motion type}
& \textbf{\#Segments}
& \textbf{Pos. err. (m)}$\downarrow$
& \textbf{Head. err. (rad)}$\downarrow$ \\
\midrule
Stationary ($v\!<\!0.1\,\mathrm{m/s}$) 
& 397488 & 0.02 & 0.00 \\
Straight ($0^{\circ}\!\leq\!|\Delta\theta|\!<\!5^{\circ}$) 
& 238393 & 0.08 & 0.01 \\
Mild turn ($5^{\circ}\!\leq\!|\Delta\theta|\!<\!22.5^{\circ}$) 
& 33713 & 0.10 & 0.03 \\
Sharp turn ($22.5^{\circ}\!\leq\!|\Delta\theta|\!<\!45^{\circ}$) 
& 11514 & 0.12 & 0.06 \\
Very sharp turn (turn-heavy) ($|\Delta\theta|\!\geq\!45^{\circ}$) 
& 860 & 0.16 & 0.19 \\
\bottomrule
\end{tabular}
}
\end{table}

\section{Additional turn-heavy and refinement analysis}
\label{sec:appendix_turn_refinement}

Table~\ref{tab:turn_refinement_effect} further analyzes the role of method-specific refinement stages on the same AV2 validation split used in Table~\ref{tab:turn_subset}. All methods are evaluated on the same 24,988 candidates, with 3,007 turn-heavy samples and 21,981 lower-turning samples. For baselines with refinement stages, we report both the pre-refinement output and the full refined output. Refinement generally improves QCNet and DONUT, confirming that post-hoc trajectory correction is beneficial for these architectures.

MoCAR does not use any refinement stage: it directly autoregresses motion codes and decodes trajectories from the latent rollout. On the turn-heavy subset, it achieves lower minFDE$_6$, minADE$_6$, and MR$_6$ than both the pre-refinement and fully refined outputs of QCNet and DONUT. These results demonstrate that MoCAR achieves strong turn-heavy forecasting accuracy with a single-stage coordinate-aware latent rollout.

\begin{table*}[h]
\centering
\caption{
Effect of method-specific refinement stages on the AV2 validation set stratified by future turning angle.
All methods are evaluated on the same 24,988 candidates and the same ground-truth turn split as Table~\ref{tab:turn_subset}, with 3,007 turn-heavy and 21,981 lower-turning samples.
Best values are in bold, and second-best values are underlined; ties are marked consistently.
}
\label{tab:turn_refinement_effect}
\resizebox{\linewidth}{!}{
\begin{tabular}{lllcccc}
\toprule
\textbf{Method}
& \textbf{Stage}
& \textbf{Subset}
& \textbf{minFDE$_6$}$\downarrow$
& \textbf{minADE$_6$}$\downarrow$
& \textbf{MR$_6$}$\downarrow$
& \textbf{b-minFDE$_6$}$\downarrow$ \\
\midrule
\multirow{6}{*}{QCNet~\citep{zhou2023query}}
& \multirow{3}{*}{w/o refinement}
& Turn-heavy ($\geq45^\circ$) & 2.40 & 1.17 & 0.38 & 2.99 \\
& & Lower-turning ($<45^\circ$) & 1.13 & \textbf{0.65} & 0.14 & 1.73 \\
& & All & 1.28 & \underline{0.72} & 0.17 & 1.88 \\
\cmidrule(lr){2-7}
& \multirow{3}{*}{w/ refinement}
& Turn-heavy ($\geq45^\circ$) & 2.37 & 1.17 & 0.36 & 2.98 \\
& & Lower-turning ($<45^\circ$) & 1.10 & \underline{0.66} & \underline{0.13} & 1.72 \\
& & All & 1.25 & \underline{0.72} & 0.16 & 1.87 \\
\midrule
\multirow{6}{*}{DONUT~\citep{knoche2025donut}}
& \multirow{3}{*}{w/o refinement}
& Turn-heavy ($\geq45^\circ$) & 2.06 & 1.11 & 0.32 & 2.68 \\
& & Lower-turning ($<45^\circ$) & \underline{1.08} & 0.68 & \textbf{0.12} & \underline{1.71} \\
& & All & 1.20 & 0.73 & 0.15 & \underline{1.82} \\
\cmidrule(lr){2-7}
& \multirow{3}{*}{w/ refinement}
& Turn-heavy ($\geq45^\circ$) & \underline{2.01} & \underline{1.10} & \underline{0.31} & \textbf{2.64} \\
& & Lower-turning ($<45^\circ$) & \textbf{1.07} & 0.67 & \textbf{0.12} & \textbf{1.70} \\
& & All & \underline{1.18} & 0.73 & \underline{0.14} & \textbf{1.81} \\
\midrule
\multirow{3}{*}{MoCAR (Ours)}
& \multirow{3}{*}{refinement-free}
& Turn-heavy ($\geq45^\circ$) & \textbf{1.99} & \textbf{1.07} & \textbf{0.28} & \underline{2.67} \\
& & Lower-turning ($<45^\circ$) & \textbf{1.07} & \textbf{0.65} & \textbf{0.12} & 1.72 \\
& & All & \textbf{1.17} & \textbf{0.70} & \textbf{0.13} & 1.84 \\
\bottomrule
\end{tabular}
}
\end{table*}

\section{Argoverse~1: preprocessing and full evaluation}
\label{sec:appendix_av1}
 
\paragraph{Zero-shot transfer protocol.}
Table~\ref{tab:av1_protocol} summarizes the changes encountered by an AV2-trained model when evaluated on AV1. The observation and prediction horizons, available motion states, and agent-label conventions differ, while the sampling rate is shared. This keeps the physical duration of a motion code fixed, allowing the unchanged AV2 checkpoint to be evaluated without temporal adaptation. For MoCAR, the shorter observed history yields fewer valid historical codes. The 3\,s future is generated in two prediction steps at the default stride. All methods use the same preprocessing and masking protocol.

\begin{table}[h]
\centering
\caption{AV2 training and AV1 zero-shot evaluation settings. A segment contains 16 points and spans 1.5\,s at the shared sampling rate.}
\label{tab:av1_protocol}
\begin{tabular}{p{0.23\linewidth}p{0.29\linewidth}p{0.37\linewidth}}
\toprule
\textbf{Property} & \textbf{AV2 training} & \textbf{AV1 zero-shot evaluation} \\
\midrule
Observation / prediction & 5\,s / 6\,s & 2\,s / 3\,s \\
Available motion states & Position, velocity, heading & Position; velocity and heading derived \\
Agent labels & Native AV2 categories & Ego and AGENT mapped to vehicle; others to other/unknown \\
Model weights & Trained on AV2 & Unchanged AV2 checkpoint; no fine-tuning \\
Sampling / code duration & 10\,Hz / 1.5\,s & 10\,Hz / 1.5\,s \\
\bottomrule
\end{tabular}
\end{table}

\paragraph{AV1 preprocessing.}
AV1 only provides agent positions; velocity and heading must be derived from position sequences. Direct finite differences on raw AV1 positions produce noisy heading estimates that perturb the endpoint-anchor frames MoCAR relies on. To reduce this noise, we apply local polynomial smoothing (Savitzky--Golay style) to each agent's observed position history, using a window of 13 and polynomial order 2. Each smoothed point is constrained to stay within $0.25\,$m of its raw position. Heading and velocity are then recomputed by finite differences on the smoothed positions. The same preprocessing is applied to all methods, including the baseline zero-shot evaluations in Table~\ref{tab:sota_av1}. All AV1 metrics are computed against the original raw future positions. Smoothing is used only to derive velocity and heading inputs. For AV1 zero-shot evaluation, the autonomous-vehicle (ego) category and AGENT category are mapped to the vehicle embedding. Other observed agents are mapped to the other/unknown embedding.
 
\paragraph{Effect of positional-noise augmentation.}
\label{sec:appendix_noise}
We train a control without the positional-noise augmentation described in Section~\ref{sec:appendix_train}, keeping all other settings unchanged. AV2 validation metrics remain unchanged at the reported precision. On AV1 zero-shot evaluation, augmentation reduces minFDE$_6$ from 1.27 to 1.24 and b-minFDE$_6$ from 1.97 to 1.93 (Table~\ref{tab:noise_augmentation}). Both variants use the same AV1 smoothing and state-derivation procedure. Position perturbations change displacement magnitudes and directions, affecting the derived velocity and heading estimates. The transfer gain is consistent with improved robustness to these motion inputs.

\begin{table}[h]
\centering
\caption{Effect of training-time positional-noise augmentation on AV2-to-AV1 zero-shot evaluation. Both models use the same AV1 preprocessing and are evaluated without noise augmentation.}
\label{tab:noise_augmentation}
\scalebox{\uniformtablescale}{
\begin{tabular}{lcccccc}
\toprule
\textbf{Noise augmentation}
& \textbf{minFDE$_1$}$\downarrow$ & \textbf{minADE$_1$}$\downarrow$
& \textbf{minFDE$_6$}$\downarrow$ & \textbf{minADE$_6$}$\downarrow$
& \textbf{MR$_6$}$\downarrow$ & \textbf{b-minFDE$_6$}$\downarrow$ \\
\midrule
Without & 5.21 & 2.40 & 1.27 & 0.73 & 0.11 & 1.97 \\
With (default) & 5.14 & 2.37 & 1.24 & 0.72 & 0.11 & 1.93 \\
\bottomrule
\end{tabular}
}
\end{table}

\paragraph{Full AV1 evaluation.}
Beyond the zero-shot comparison in the main text, we evaluate two additional MoCAR configurations on AV1: short fine-tuning of the AV2-trained model for 10 epochs, and full training on AV1 from scratch. Table~\ref{tab:sota_av1_full} reports the complete results alongside published AV1-trained baselines. Fine-tuning on all agents reaches 1.03 minFDE$_6$, fine-tuning on focal and AV agents reaches 0.92, and training from scratch reaches 0.91, competitive with AV1-specialized methods and within 0.04 minFDE of the best listed result. Together with the zero-shot results, this shows that the AV2-trained MoCAR model is usable directly, substantially improves with brief fine-tuning, and remains competitive when trained on AV1 from scratch.
 
\begin{table}[h]
\centering
\caption{Full Argoverse~1 validation evaluation. Top block: representative methods originally trained on AV1 (numbers from their original papers). Middle block: zero-shot transfer of AV2-trained models, repeated from Table~\ref{tab:sota_av1} for completeness. Bottom block: MoCAR under fine-tuning and full AV1 training.}
\label{tab:sota_av1_full}
\scalebox{\uniformtablescale}{
\begin{tabular}{l|ccc}
\toprule
\textbf{Method}
& \textbf{minFDE$_6$}$\downarrow$
& \textbf{minADE$_6$}$\downarrow$
& \textbf{MR$_6$}$\downarrow$ \\
\midrule
TNT~\citep{zhao2021tnt} & 1.29 & 0.73 & 0.09 \\
LaneRCNN~\citep{zeng2021lanercnn} & 1.19 & 0.77 & 0.08 \\
mmTransformer~\citep{liu2021multimodal} & 1.15 & 0.71 & 0.11 \\
LaneGCN~\citep{liang2020learning} & 1.08 & 0.71 & -- \\
DenseTNT~\citep{gu2021densetnt} & 1.05 & 0.73 & 0.10 \\
SSL-Lanes~\citep{bhattacharyya2023ssl} & 1.01 & 0.70 & 0.09 \\
FRM~\citep{park2023leveraging} & 0.99 & 0.68 & -- \\
HiVT~\citep{zhou2022hivt} & 0.96 & 0.66 & 0.09 \\
ADAPT~\citep{aydemir2023adapt} & 0.95 & 0.67 & 0.08 \\
LAformer~\citep{liu2024laformer} & 0.92 & 0.64 & -- \\
DeMo~\citep{zhang2024decoupling} & 0.90 & \textbf{0.59} & 0.07 \\
HPNet~\citep{tang2024hpnet} & \textbf{0.87} & 0.64 & \textbf{0.07} \\
\midrule
\multicolumn{4}{l}{\textit{Zero-shot transfer of AV2-trained models:}} \\
DeMo (zero-shot) & 2.03 & 1.08 & 0.25 \\
DONUT (zero-shot) & 1.58 & 0.97 & 0.19 \\
QCNet (zero-shot) & 1.55 & 0.95 & 0.19 \\
MoCAR (zero-shot) & 1.24 & 0.72 & 0.11 \\
\midrule
\multicolumn{4}{l}{\textit{MoCAR: fine-tuning and AV1 training:}} \\
MoCAR fine-tuned, all agents (10 ep.) & 1.03 & 0.64 & 0.10 \\
MoCAR fine-tuned, focal+AV (10 ep.) & 0.92 & 0.62 & 0.08 \\
MoCAR trained on AV1 (from scratch) & 0.91 & 0.62 & 0.08 \\
\bottomrule
\end{tabular}
}
\end{table}

\section{Qualitative results}
\label{sec:appendix_qual}

\begin{figure}[t]
    \centering
    \includegraphics[width=\linewidth]{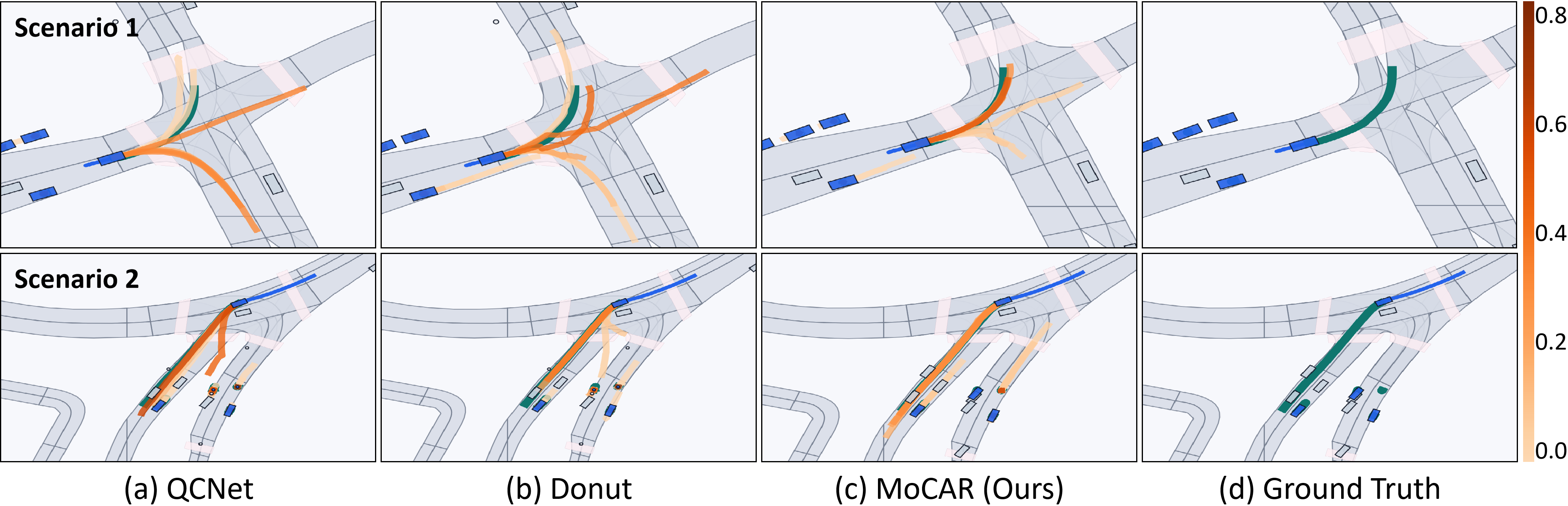}
    \caption{
    \textbf{Qualitative comparison on challenging AV2 scenarios.}
    From left to right, we show predictions from QCNet, DONUT, MoCAR, and the ground truth.
    Blue denotes the observed history, orange denotes predicted future trajectories, and green denotes the ground-truth future.
    While all methods are generally map-aligned, MoCAR produces more coherent and temporally consistent predictions in complex intersections and interaction-heavy scenes.
    }
    \label{fig:qualitative}
\end{figure}

\paragraph{Comparison with other state-of-the-art methods.}
Figure~\ref{fig:qualitative} compares QCNet, DONUT, and MoCAR on challenging AV2 scenarios. All methods remain broadly map-aligned in regular cases. The differences emerge in large intersections and interaction-heavy scenes, where forecasting requires resolving maneuver ambiguity while preserving temporal consistency. In these scenarios, QCNet and DONUT become more dispersed or less coherent after the turn, whereas MoCAR follows the feasible path more consistently. This pattern is consistent with latent rollout preserving cross-segment consistency without trajectory-space re-tokenization.

\begin{figure*}[h]
    \centering
    \includegraphics[width=\linewidth]{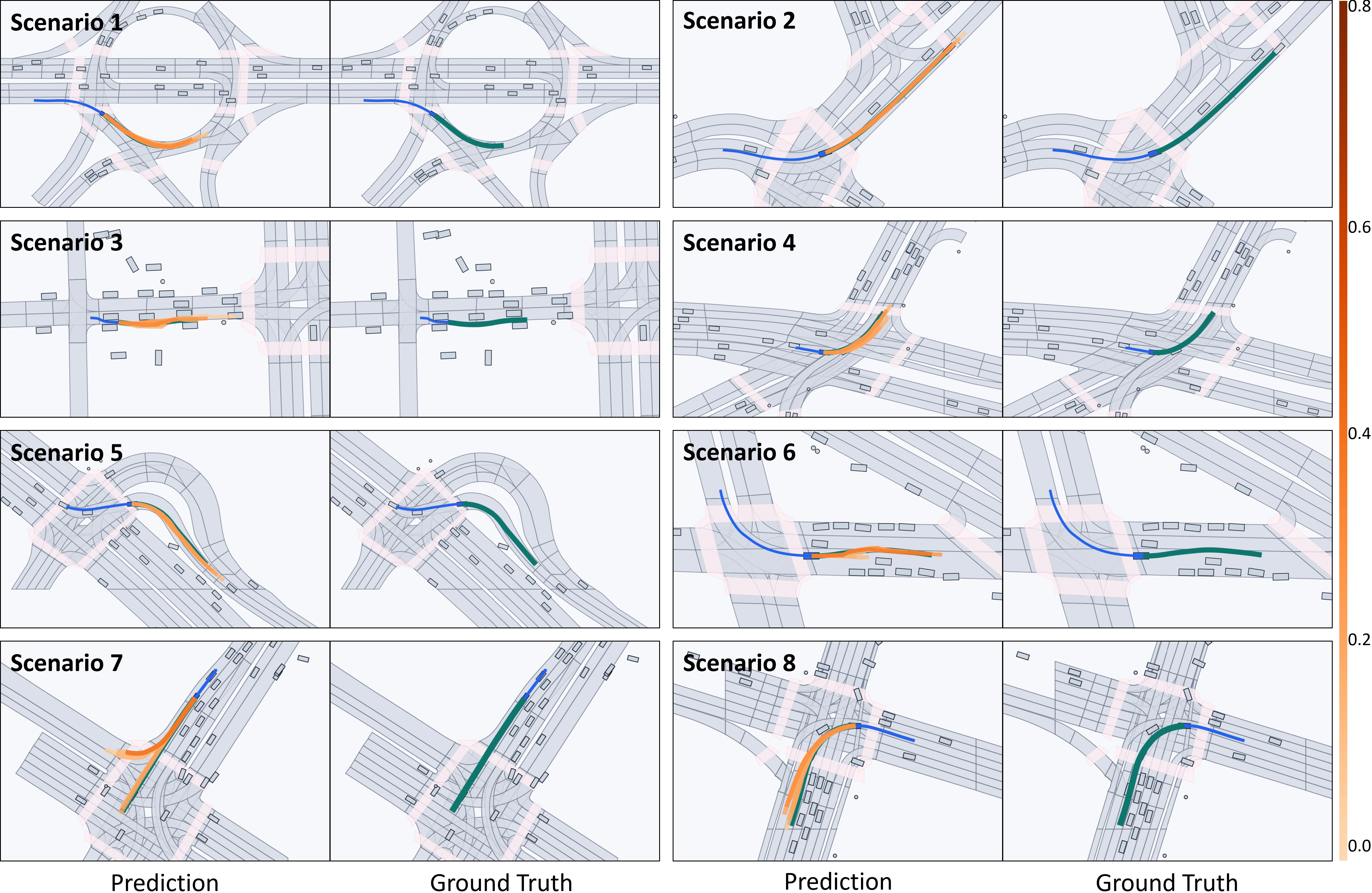}
    \caption{
    \textbf{Additional qualitative success cases on the AV2 validation set.}
    Eight representative examples covering roundabouts, straight driving, curved-road following, and turning maneuvers at intersections.
    Blue shows the observed history, green the ground-truth future, and orange the predicted futures, with darker shades indicating higher predicted probability.
    The predictions remain consistent with the local map structure and closely follow the ground-truth motion.
    }
    \label{fig:more_success}
\end{figure*}

\paragraph{Additional success cases.}
Figure~\ref{fig:more_success} shows additional qualitative examples on the AV2 validation set. Across diverse scenarios, the model produces predictions that remain well aligned with the local road geometry and closely match the ground-truth future. In addition, the supplementary material provides dynamic \textbf{video visualizations} that show the continuous forecasting results over time, further demonstrating the temporal consistency and stability of the proposed model.

\begin{figure*}[h]
    \centering
    \includegraphics[width=\linewidth]{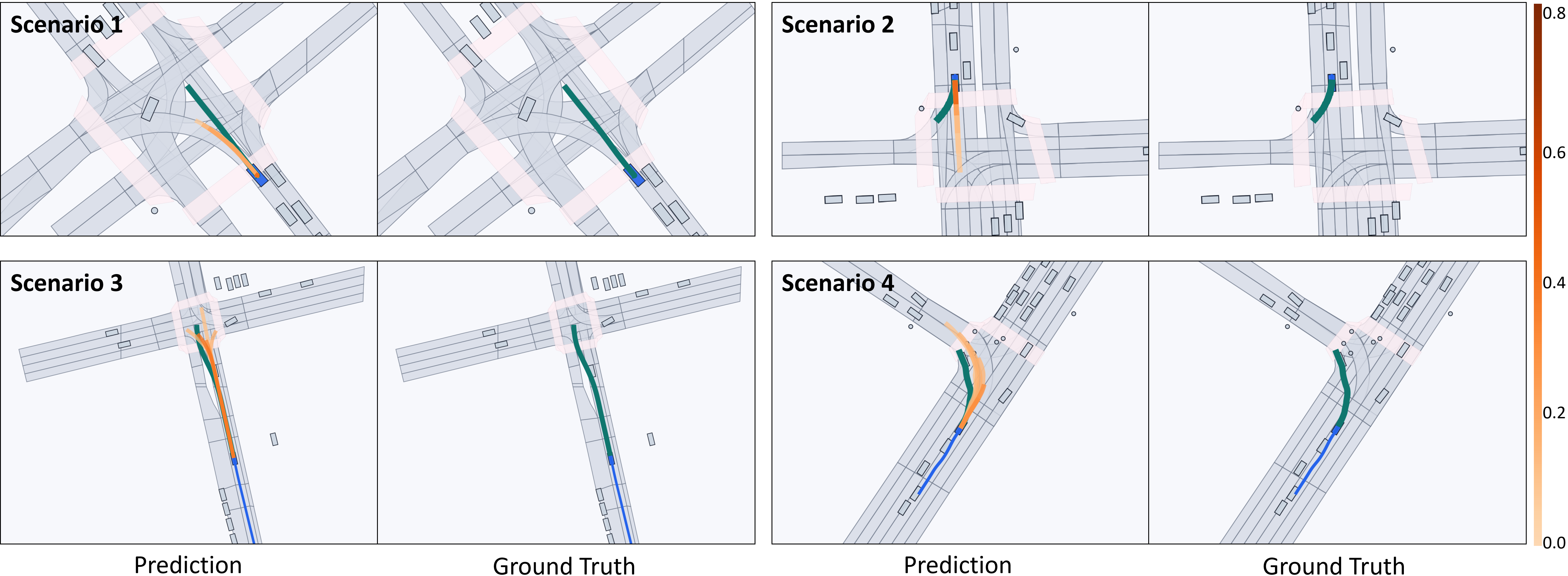}
    \caption{
    \textbf{Representative failure cases on the AV2 validation set.}
    Blue shows the observed history, green the ground-truth future, and orange the predicted futures, with darker shades indicating higher predicted probability.
    Most errors occur when the model follows the most map-consistent continuation, but the realized behavior reflects a later or less common maneuver.
    }
    \label{fig:failures}
\end{figure*}

\paragraph{Failure case analysis.}
We manually inspected representative cases from the AV2 validation set where MoCAR produces large displacement errors. As shown in Figure~\ref{fig:failures}, the illustrated failures involve a mismatch between map-consistent prediction and the realized driving behavior. MoCAR tends to favor trajectories aligned with local lane geometry and scene topology. This preference generally helps, but can lead to errors when an agent makes a late directional commitment or executes a less common maneuver. The model can also miss a continuation weakly supported by the local map structure. In these cases, it predicts a topologically plausible continuation, while the ground-truth future reflects a rarer or less predictable behavioral choice. These examples highlight the challenge of handling behavioral deviations and rare decisions that are difficult to infer from history and map context alone.

\section{Broader Impacts}
\label{sec:broader_impacts}

MoCAR is evaluated as an offline trajectory-forecasting component on public autonomous-driving benchmarks. More accurate and efficient forecasting may benefit downstream planning research, especially in complex maneuvers and cross-dataset transfer settings.
Beyond road-agent forecasting, the decoder-only motion-code formulation may also inform related 2D and 3D motion prediction problems where agents move under evolving local coordinate frames, such as maritime navigation, UAV trajectory prediction, and robotic motion forecasting.
However, incorrect forecasts could be unsafe if used directly in deployed systems. Real-world deployment would require extensive safety validation, uncertainty handling, monitoring under distribution shift, and integration with fail-safe planning and control modules.

\section{Asset usage and license compliance}
\label{app:license}

\paragraph{Datasets and benchmarks.}
We use Argoverse 2~\citep{wilson2023argoverse} and Argoverse 1~\citep{chang2019argoverse} following their official terms of use. 
Both datasets are public autonomous-driving forecasting benchmarks released under the Creative Commons Attribution-NonCommercial-ShareAlike 4.0 International license (CC BY-NC-SA 4.0). 
The associated Argoverse code and APIs are released under the MIT license.

\paragraph{External code and model assets.}
For baseline evaluation, we use publicly released implementations and pretrained checkpoints of QCNet~\citep{zhou2023query}, DONUT~\citep{knoche2025donut}, and DeMo~\citep{zhang2024decoupling}. 
The official QCNet repository is released under the Apache-2.0 license and provides pretrained checkpoints for Argoverse 2 evaluation. 
The official DONUT repository is also released under the Apache-2.0 license, builds upon QCNet, and provides a pretrained checkpoint for evaluation. 
The official DeMo repository provides publicly available checkpoints; however, we did not find an explicit repository-level license for its code or checkpoints. 
We therefore use DeMo only for academic benchmarking and comparison, with proper citation, and do not redistribute its code, checkpoints, or derived model assets.

\paragraph{Compliance.}
All external assets are used only for academic research, benchmarking, and reproducibility. 
The supplementary material does not include third-party datasets, external codebases, pretrained checkpoints, or non-redistributable assets. 
The MoCAR implementation and AV2 preprocessing, training, validation, and visualization scripts are publicly available at \url{https://github.com/XuYiMing83/MoCAR}. The repository includes installation and usage instructions and an MIT license. Model checkpoints will be released upon publication. The release will preserve the required attribution and will not redistribute external assets beyond their original license terms.

% \clearpage

%%%%%%%%%%%%%%%%%%%%%%%%%%%%%%%%%%%%%%%%%%%%%%%%%%%%%%%%%%%%
% \newpage
% \input{checklist_camera_ready.tex}

\end{document}